\documentclass[letterpaper, 10 pt, journal, twoside]{IEEEtran}
\IEEEoverridecommandlockouts                              

\usepackage{subcaption}
\usepackage{mwe}
\usepackage{graphics} 
\usepackage{bm}
\usepackage{siunitx}
\usepackage{relsize}
\usepackage{amsmath} 
\usepackage{amssymb}  
\usepackage{booktabs}
\usepackage{siunitx}
\usepackage{multirow}
\usepackage{makecell}
\usepackage{subcaption} 
\usepackage{hyperref}
\usepackage[usenames, dvipsnames,table,xcdraw]{xcolor}
\usepackage{pifont}

\usepackage[capitalise]{cleveref}
\usepackage{bm}
\usepackage{mathtools}
\usepackage{cite}

\newcommand{\rsec}                                      [1]     {Section~\ref{#1}}
\renewcommand{\cdot}{}

\usepackage[firstpage]{draftwatermark}
\SetWatermarkText{\begin{tabular}{l}
    \copyright 2023 IEEE.  Personal use of this material is permitted. 
    Permission from IEEE must be obtained for all other uses$\textrm{,}$ in any current or future media$\textrm{,}$ including reprinting/republishing this material\\ 
    for advertising or promotional purposes$\textrm{,}$ creating 
    new collective works$\textrm{,}$ for resale or redistribution to servers or lists$\textrm{,}$ or reuse of any copyrighted component of this work in other works.
    \end{tabular}}
\SetWatermarkColor[gray]{0}
\SetWatermarkFontSize{0.22cm}
\SetWatermarkVerCenter{0.5cm}
\SetWatermarkAngle{0}
\SetWatermarkHorCenter{10.13cm}
\usepackage[english]{babel}
\addto\extrasenglish{%
}

\newcolumntype{P}[1]{>{\centering\arraybackslash}p{#1}}
\newcolumntype{M}[1]{>{\centering\arraybackslash}m{#1}}

\definecolor{changedcol}{rgb}{0.,0.,0.}
\DeclareCaptionFont{ccchngd}{\color{changedcol}}
\DeclareCaptionFont{ccblack}{\color{black}}

\newenvironment{Changed}{%
\captionsetup{textfont={ccchngd},labelfont={ccchngd}}%
\color{changedcol}%
}%
{\captionsetup{textfont={ccblack},labelfont={ccblack}}}
\usepackage{todonotes}
\usepackage{fixfoot}
\DeclareFixedFootnote{\githubLink}{\href{https://github.com/dfki-ric-underactuated-lab/acromonk}{https://github.com/dfki-ric-underactuated-lab/acromonk}}
\DeclareFixedFootnote{\videoLink}{\href{https://youtu.be/FIcDNtJo9Jc}{https://youtu.be/FIcDNtJo9Jc}}
\DeclareFixedFootnote{\OpensourceLink}{\href{https://youtu.be/FIcDNtJo9Jc}{https://youtu.be/FIcDNtJo9Jc}}

\title{AcroMonk: A Minimalist Underactuated Brachiating Robot}

\author{Mahdi Javadi, Daniel Harnack, Paula Stocco, Shivesh Kumar, Shubham Vyas, \\Daniel Pizzutilo, and Frank Kirchner
\thanks{Manuscript received: November 29, 2022; Revised March 3, 2023; Accepted April 4, 2023.}
\thanks{This paper was recommended for publication by Editor L. Pallottino upon evaluation of the Reviewers' comments.
This work has been supported by the M-RoCK~(FKZ 01IW21002) and VeryHuman~(FKZ 01IW20004) projects 
funded by the German Aerospace Center (DLR) with federal funds from the Federal Ministry of 
Education and Research (BMBF) and is additionally
supported with project funds from the federal state of Bremen for setting up
the Underactuated Robotics Lab (201-342-04-2/2021-4-1). 
The fifth author acknowledges support from the Stardust Reloaded project which 
has received funding from the European Union’s 
Horizon 2020 research and innovation program under the 
Marie Skłodowska-Curie grant agreement No 813644. (\textit{Corresponding author: Mahdi Javadi})} 
\thanks{Mahdi Javadi, Daniel Harnack, Shivesh Kumar, Shubham Vyas, Daniel Pizzutilo, and Frank Kirchner are with Robotics Innovation Center, DFKI GmbH, 28359 Bremen, Germany. 
(e-mail: Mahdi.Javadi@dfki.de; Daniel.Harnack@dfki.de; Shivesh.Kumar@dfki.de; Shubham.Vyas@dfki.de; Daniel.Pizzutilo@dfki.de; Frank.Kirchner@dfki.de)}%
\thanks{Paula Stocco is with Department of Mechanical Engineering, Stanford University, CA, USA. (e-mail: stoccop@stanford.edu)}%
\thanks{Shubham Vyas and Frank Kirchner are also with AG Robotik, University of Bremen, 28359 Bremen, Germany.}%
\thanks{Digital Object Identifier (DOI): \href{https://ieeexplore.ieee.org/document/10106397}{10.1109/LRA.2023.3269296}}     
\thanks{}
}

\begin{document}
\newcommand{\robotName}{AcoMonk}

\newcommand{\mvec}[1]{\bm{#1}}
\newcommand{\vc}[1]{\mathbf{\mathbf{#1}}}

\newcommand{\q}{\textbf{q}}
\newcommand{\dq}{\dot{\q}}
\newcommand{\ddq}{\ddot{\q}}


\newcommand{\Mass}{\mathbf{M}}
\newcommand{\Bias}{\mathbf{b}}
\newcommand{\Gravity}{\mathbf{g}}
\newcommand{\Force}{\mathbf{\lambda}}
\newcommand{\Torque}{\mathbf{\tau}}
\newcommand{\Jac}{\mathbf{J}}

\newcommand{\BIN}{\begin{bmatrix}}
\newcommand{\BOUT}{\end{bmatrix}}

\newcommand{\sref}[1]{Sec~\ref{#1}}
\newcommand{\eref}[1]{(\ref{#1})}
\newcommand{\fref}[1]{Fig.~\ref{#1}}
\newcommand{\tref}[1]{Table~\ref{#1}}
\newcommand{\qref}[1]{(Eq.\ref{#1})}
\newcommand{\state}{\mathbf{x}}
\newcommand{\ctrl}{\mathbf{u}}
\newcommand{\dynsys}{\mathbf{f}}

\newcommand{\qTr}{\underline{\q}}
\newcommand{\dqTr}{\underline{\dq}}
\newcommand{\ddqTr}{\underline{\ddq}}
\newcommand{\TorqueTr}{\underline{\Torque}}

\newcommand{\costl}{l}
\newcommand{\dts}{\Delta t_s}
\newcommand{\st}{\text{subject to}}

\maketitle

\begin{abstract}
Brachiation is a dynamic, coordinated swinging maneuver of body and arms used by monkeys and apes to move between branches. 
As a unique underactuated mode of locomotion, it is interesting to study from a robotics perspective since it can broaden the deployment scenarios for humanoids and animaloids. 
While several brachiating robots of varying complexity have been proposed in the past, this paper presents the simplest possible prototype of a brachiation robot, using only a single actuator and unactuated grippers. 
The novel passive gripper design allows it to snap on and release from monkey bars, while guaranteeing well defined start and end poses of the swing. 
The brachiation behavior is realized in three different ways, using
trajectory optimization via direct collocation and stabilization by a model-based time-varying linear quadratic regulator (TVLQR) or model-free proportional derivative (PD) control, as well as by a reinforcement learning (RL) based control policy. 
The three control schemes are compared in terms of robustness to disturbances, mass uncertainty, and energy consumption. 
The system design and controllers have been open-sourced\footnote{The open-source implementation is available 
at \href{https://github.com/dfki-ric-underactuated-lab/acromonk}{https://github.com/dfki-ric-underactuated-lab/acromonk} and a video demonstration of the experiments can be accessed at \href{https://youtu.be/FIcDNtJo9Jc}{https://youtu.be/FIcDNtJo9Jc}.}. 
Due to its minimal and open design, the system can serve as a canonical underactuated platform for education and research. 
\end{abstract}
\IEEEpeerreviewmaketitle
\newcommand{\sectionspace}{-0.3cm}
\vspace{\sectionspace}
\section{Introduction}
\label{sec:introduction}
\IEEEPARstart{B}{rachiation} is a complex dynamic maneuver involving a continuous swing motion and a discontinuity when switching the support arm. 
Apes brachiate with ease through unstructured environments with flexible or rigid handholds at variable distances, making this motion challenging and interesting to study for roboticists. 
Brachiating robots can be beneficial for inspection, agriculture, search and rescue applications, etc., since they can perform agile movements in hard to traverse terrains. 
Hence, there has been extensive research on brachiation robots in the past three decades. 

\begin{figure}[t]
    \centering
    \includegraphics[width=0.83\linewidth]{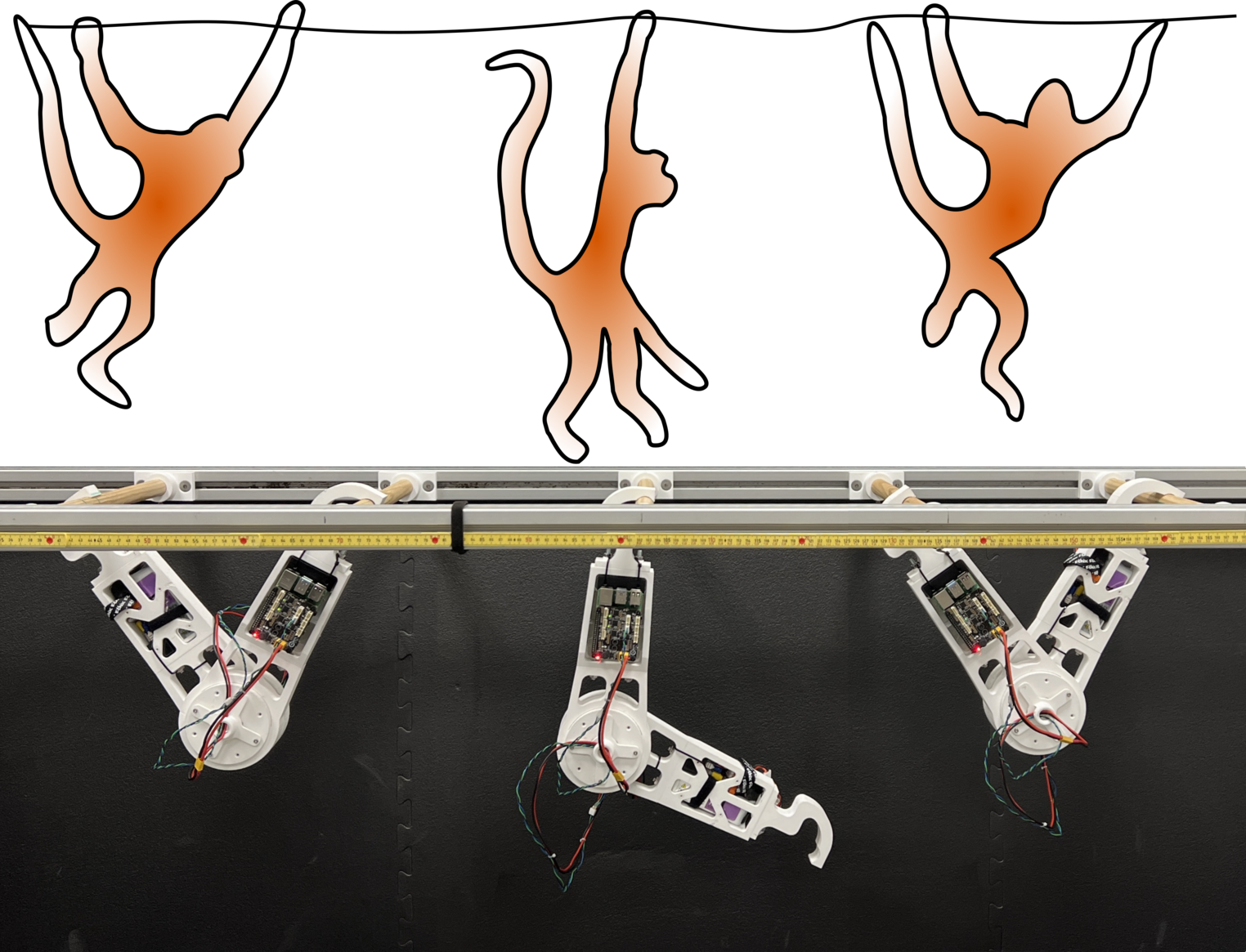}
    \caption{Monkey inspired brachiation with AcroMonk}
    \label{fig:monkey-brachiation}
    \vspace{-0.7cm}
\end{figure}

One of the first brachiating mobile robots, Brachiator I~\cite{fukuda-first}, was introduced in 1991 by Fukuda et al., consisting of six links and five joints.
Many designs followed this seminal work, ranging from simplified 
systems of two joints and actuated grippers~\cite{fukuda-swing, meghdari2013minimum, davies2018tarzan, farzan-tvlqr} that could traverse 
rigid bars and flexible ropes, over more complex systems with a passive tail~\cite{yang2019design} for stabilization and a conceptual seven link design~\cite{hasegawa1999motion}, to a full ape-like robot with 12 joints and active grippers that allowed the system to perform realistic, monkey-like swings~\cite{fukuda_brachiator3}.
Realizing different brachiation types, i.e. ladder, rope brachiation or ricocheting, comprises different challenges and enforces distinct demands for 
behavior generation and control. 
Formulating the behavior generation as a trajectory optimization problem provides flexibility to incorporate desired demands in terms of costs and constraints.
A desired trajectory can thus be generated by employing mathematical approaches using the robot's physical parameters and the grasping configuration~\cite{fukuda-first}~\cite{hook-shaped}. 
Furthermore, introducing the system's mechanical energy in the problem formulation allows optimizing trajectories 
by using physical energy conservation.
These trajectories can be generated offline 
and be stabilized during execution to achieve the 
desired behavior on the robot~\cite{meghdari2013minimum}~\cite{yang2019design}~\cite{farzan-trajopt}.
Stabilization requires an online controller taking state feedback into account. 
Among the various controllers employed for this purpose, 
PD controllers are most commonly used~\cite{hasegawa2000behavior}~\cite{saito1994learning}~\cite{hook-shaped}. 
Machine learning and heuristic methods are also popular, providing a model-free approach to learn brachiation behavior~\cite{fukuda-heuristic}~\cite{saito1994learning}~\cite{cheng2018deep}. 
Most recent research however focuses on model-based and energy-optimal control approaches, since energy-optimal formulations bring the advantage of robustness against uncertainties and can be incorporated in behavior generation~\cite{farzan-trajopt}~\cite{meghdari2013minimum}, behavior control~\cite{farzan-sos}, or both~\cite{farzan-tvlqr}. 
Table~\ref{tab:sota} summarizes the brachiation robot literature 
in categories of system design, behavior generation, and control approaches. 
\vspace{-0.1cm}
\begin{table}[!htpb]
	\centering
	\scriptsize
	\caption{Overview of brachiation robots. (L, J, A, G) indicate the number of links, joints and actuators, and type of grippers respectively.}
	\begin{center}
			\vspace{-0.1cm}
		\begin{tabular}{ccc}
			\toprule
			Category field& Description & Reference(s) \\
			\midrule
			\multirow{5}{*}{System Design
				(L, J, A, G)}
			&(2, 1, 1,  Active)&\scriptsize\cite{fukuda1991study}\cite{fukuda-heuristic}\cite{fukuda-target-dynamics}
			\\
			&(3, 2, 2, Active)   & \cite{yang2019design}\\         
			&(7, 6, 6, Active)   &\cite{fukuda2013modification}\cite{hasegawa2001behavior}\cite{nakanishi1999brachiation}
			\\
			&(13, 12, 14, Active)   &\cite{fukuda2013modification}\cite{hasegawa1999motion}\cite{nakanishi1999brachiation}
			\\        
			&(2, 2, 2, Passive)   & \cite{hook-shaped}\cite{hook-shaped-second}\\
			\midrule
			\multirow{4}{*}{Trajectory Generation}
			&Heuristic Methods &\cite{fukuda-heuristic}\cite{fukuda-target-dynamics}\cite{fukuda-swing}
			\\
			&Harmonic Oscillator   &\cite{fukuda-target-dynamics}\cite{nakanishi2000leaping}\\
			&Energy minimization   &\cite{farzan-trajopt}\cite{farzan-tvlqr}\cite{farzan-sos}\cite{meghdari2013minimum}\\
			&Posture-based   & \cite{hook-shaped}\cite{hook-shaped-second}\\  
			\midrule
			\multirow{5}{*}{Behavior Control}
			&Machine learning &\cite{hasegawa1999self}\cite{fukuda-heuristic}\cite{saito1994learning}
			\\
			&PD controller &  \cite{fukuda-heuristic}\cite{hasegawa1999self}\cite{meghdari2013minimum}
			\\  
			
			&Model Predictive Control   &\cite{oliveira-mpc}\cite{oliveira-nmpc}\cite{oliveira-realtime}\\
			&Input/output linearization   &\cite{nakanishi1997preliminary}\cite{spong-pfl}\cite{yang2019design}\\
			&Energy based controller   &\cite{farzan-trajopt}\cite{farzan-tvlqr}\cite{farzan-sos}\\
			\bottomrule
		\end{tabular}
	\end{center}
	\vspace{-0.5cm}
	\label{tab:sota}
\end{table}


%

While several brachiating robots of varying complexity, 
along with a range 
of control strategies, 
have already been proposed, most robots include active grippers which
leads to a complex system design prone to high maintenance and electro-mechanical failure points.  
The only system with passive grippers proposed so far~\cite{hook-shaped}~\cite{hook-shaped-second} is fully actuated with two motors and was not able to execute more than two continuous brachiation maneuvers. Thus, 
there is a lack of a robust minimalist system which allows the study of underactuated brachiation.
To fill this gap, we propose AcroMonk, a novel underactuated brachiation robot with a single motor (see Figure~\ref{fig:monkey-brachiation}). A quasi-direct drive (QDD) is chosen as the actuator with a gear ratio of 6:1 which offers low friction and high backdriveability essential for dynamic locomotion. 
Its unique passive grippers feature a double grooved design, which results in a large region of attraction for grasping a target bar and a well defined rotation point for swing maneuvers. We show that AcroMonk is able to robustly brachiate continuously over a horizontal ladder with a wide range of controller types, using direct collocation for trajectory optimization and trajectory stabilization, either with model-based TVLQR or model-free PD control, or a RL-based policy. 
All three control methods are compared in terms of robustness against disturbances, modeling inaccuracies, and energy consumption. 
The simplicity of the robot's design, low maintenance requirements, 
and ease of controllability makes it a suitable platform for underactuated robotics education and research. 
The platform has been open-sourced\githubLink
~(in the spirit of~\cite{Wiebe2022, 2022_rss_realaigym, 2023_wiebe_doublependulum}), to encourage its use in research and education.
 The performance of the AcroMonk in hardware tests is shown in the accompanying video\videoLink. 
\paragraph*{Organization}
\rsec{sec:mechatronics} outlines the mechatronics system design of the AcroMonk robot. 
\rsec{sec:trajopt} addresses behavior generation methods using trajectory optimization and RL. \rsec{sec:control} details the behavior controllers for the robot and 
\rsec{sec:results} the controller comparison results in hardware experiments.
Finally, \rsec{sec:conclusion} concludes the paper and addresses future research directions.



\vspace{\sectionspace}
\section{Mechatronics System Design}
\label{sec:mechatronics}
\vspace{-0.05cm}
The motivation of the mechatronic system design of the AcroMonk was to achieve a minimalist system to study dynamic brachiation.
Additionally, we aimed for a compact design which fits in a backpack and 
can be operated as a self-sustained system for classroom teaching. 
\vspace{-0.15cm}

\begin{figure}[t]
    \centering
    \includegraphics[width=0.9\linewidth]{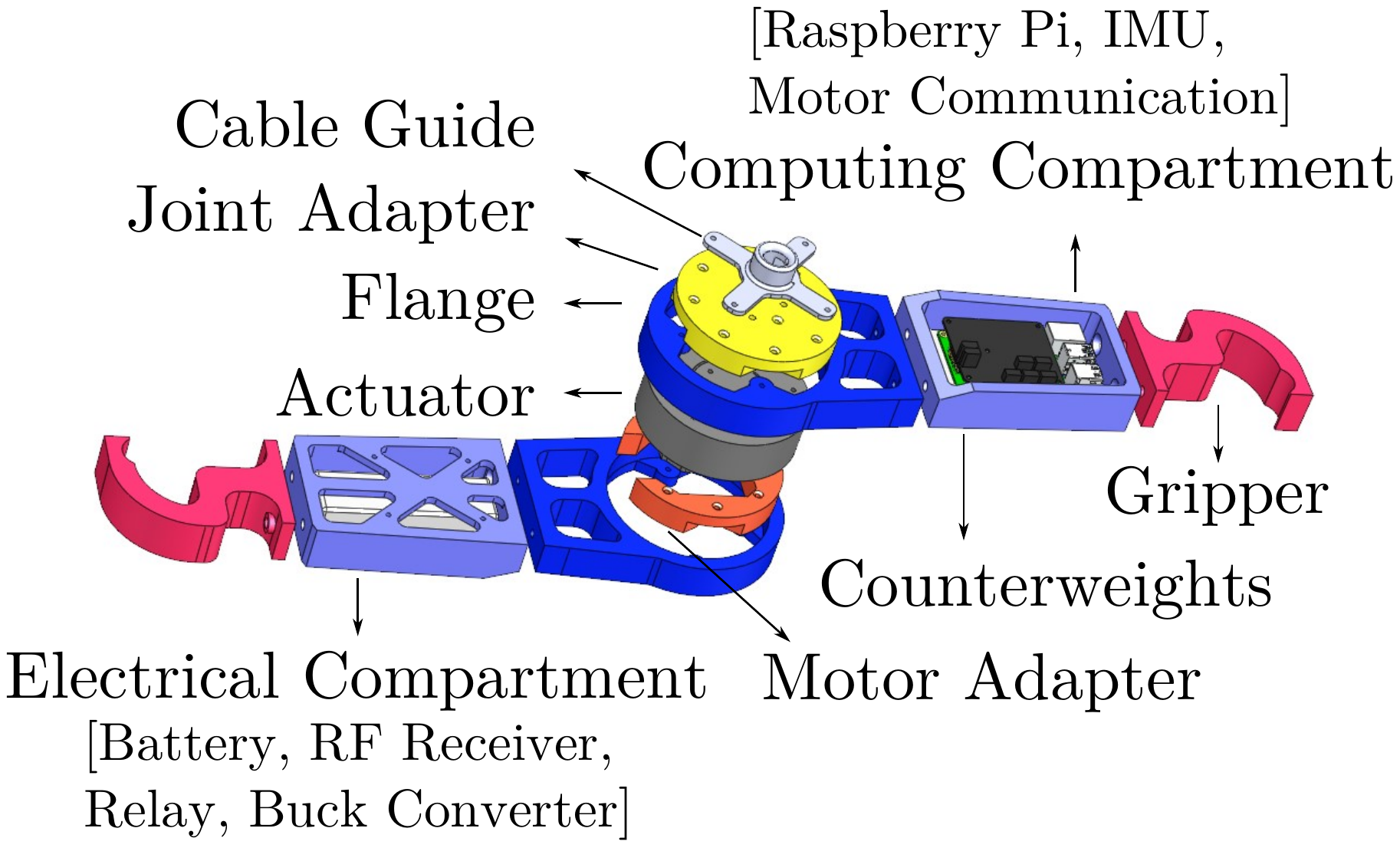}
    \caption{Mechatronic system design of AcroMonk}
    \label{fig:acromonk-design}
    \vspace{-0.7cm}
\end{figure}

\vspace{-0.2cm}
\subsection{Mechanical Design}
\vspace{-0.1cm}
The mechanical design choices were guided by using readily available hardware 
for ease of reproducibility and achieving a structure that is robust 
to falls and easy to repair.
These goals led to a modular design with one central motor connecting 
two arms that can be 3D printed with readily available materials (BASF Ultrafuse PLA).
Overall, the structure consists of six unique 3D-printed parts highlighted with different colors in~\autoref{fig:acromonk-design}, connected by screw-nut fasteners for easy assembly, with compartments for electronics, a battery, counterweights, and cable guides. 
Computing and electrical equipment are mounted on opposite arms to ensure an even mass-inertia distribution between the arms. 
For continuous brachiation, special deliberation was given to the gripper design. 
The gripper should provide sufficient error tolerance for grasping during the brachiating maneuvers while providing a defined rotation point for the next swing once connected to the bar. 

This was realized by a relatively wide opening angle of the hook, an incline towards a groove where the hook comes to rest, and an off-center connection to the arm. As illustrated in Figure 4, the intentional misalignment of the gripper’s stable point aids in sliding towards the groove. 
The slope of the inclined surface is chosen through empirical observations as 20~degrees for angle of attack with overall radius of 35~mm. These values depend on the friction coefficient of the material pairing of the gripper surface (PLA) and monkey bars (wood) and normal force.  
A higher friction coefficient implies a steeper angle to ensure slipping into the groove with minimal wobbling. 
Consequently, within the expected deviations from an ideal movement, the hook comes to rest in the groove, providing a defined rotation point for next brachiation. 
\vspace{-0.2cm}
\subsection{Electrical and Processing Architecture}
\vspace{-0.1cm}
For the actuator, the mjbots qdd100 Quasi-Direct Drive with a gear ratio of 6:1, a maximum speed of 40 rad/s, maximum continuous torque of 6 Nm, and a peak torque of 16 Nm was used. 
A Raspberry Pi 4 mounted in the computing compartment was selected as an on-board control computer due to its small form factor. 
The add-on board pi3hat for Raspberry Pi from mjbots was used to communicate with the motor via the Controller Area Network (CAN) bus. It includes an Inertial Measurement Unit (IMU) for state estimation. 
Due to the single motor design, only the relative angle between the links can be directly measured. 
The angle and angular velocity of the support arm with respect to the vertical axis were computed using the IMU, 
resulting in a full state feedback of the system. 
The computing setup allows for real-time position, velocity, and torque control at a maximum frequency of 300 Hz with Python3. 
All electronics are powered by a 6S 1200 mAh LiPo battery. 
For safety, a wireless emergency stop was implemented using a hobby-grade radio control (RC) remote and receiver combined with a direct current (DC)-DC converter and a relay switch. 

\vspace{\sectionspace}
\section{Behavior Generation}
\label{sec:trajopt}
Assuming that one support arm is always in contact with a bar, AcroMonk has two independent degrees of freedom (DOF) 
with one passive DOF at the shoulder ($q_1$) and one active DOF at the elbow ($q_2$). 
	Let $\mathbf{q}={[q_1,q_2]}^T\in\mathbb{R}^2$, 
	$\mathbf{\dot{q}}={[\dot{q}_1,\dot{q}_2]}^T\in\mathbb{R}^2$, 
	$\mathbf{\ddot{q}}={[\ddot{q}_1,\ddot{q}_2]}^T\in\mathbb{R}^2$ denote the generalized positions, velocities and accelerations. 
Its system dynamics is similar to acrobot~\cite{Spong1995} and is given by:
\vspace{-0.07cm}
\begin{equation}
	\label{eq:eom}
	\mathbf{M}(\mathbf{q}) \ddot{\mathbf{q}} +\mathbf{C}(\mathbf{q}, \dot{\mathbf{q}})\dot{\mathbf{q}} = \mathbf{\tau}_g (\mathbf{q}) + \mathbf{B} u
	\vspace{-0.07cm}
\end{equation} 
where $\mathbf{M}(\mathbf{q})$ denotes the mass-inertia matrix, $\mathbf{C}(\mathbf{q}, \dot{\mathbf{q}})$ denotes the Coriolis and centrifugal matrix, $\mathbf{\tau}_g(\mathbf{q})$ comprises the gravity effects, 
the actuation matrix is $\mathbf{B} = [0 \quad 1]^T$, and $u \in \mathbb{R}$ is the motor torque.
The AcroMonk's schematic with the base and end-effector points is depicted in Figure~\ref{fig:2darm:coordinate}. 

	Different colors are used to distinguish the support arm (blue) and the swing arm (green).

Inspired by the typical brachiation of a monkey depicted in \autoref{fig:monkey-brachiation}, we define four atomic sub-behaviors, the sequential composition of which can give rise to robust bidirectional brachiation over horizontal bars. 
In the following, we discuss the behavior state machine and methods to generate atomic behaviors including releases, swings, and grasps. For a better understanding, also refer to corresponding sections of the accompanying video.  
%
%
\vspace{-0.3cm}

	{
		\begin{figure}[!htbp]
			\centering
			\includegraphics[width=1.0\linewidth]{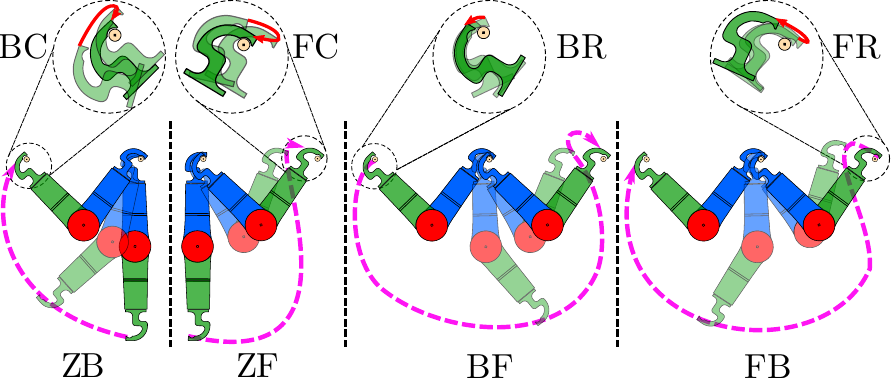}	
			\caption{Visualization of the behavior state machine}
			\label{fig:statemachine}
		\end{figure}		
	}

\vspace{-0.8cm}
\subsection{Behavior State Machine}

Considering a system that comprises the robot and bars, we denote three fixed points as Z (single support, hanging), B (double support with swing arm on backward bar), and F (double support with swing arm on forward bar). 
The four atomic sub-behaviors are transitions between these fixed points, i.e. Zero-to-Back (ZB), Zero-to-Front (ZF), Front-to-Back (FB), and Back-to-Front (BF).
Because of the passive gripper choice, additional behaviors have to be considered to release the swing arm from a bar, which is denoted as Back Release (BR) and Front Release (FR) to initiate a BF or FB atomic behavior, respectively. To ensure that the hook rests in the groove before changing the support arm, Front Catch (FC) and Back Catch (BC) are necessary for grasping the bar from above and below.
Schematic evolutions of the BR and FR motions are depicted in Figure~\ref{fig:statemachine}, where the arrows illustrate the frame progression.
Forward and backward brachiations result from a given sequence of the described swing and gripper behaviors. Finally, ZF and ZB transitions can either serve as the starting or recovery phase.
As an example, consider the sequence ZB~$\rightarrow$~BC~$\rightarrow$~BR~$\rightarrow$~BF~$\rightarrow$~FC~$\rightarrow$~BR~$\rightarrow$~BF~$\rightarrow$~FC resulting in two forward brachiation (not to be confused with FB) maneuvers starting from zero configuration of the robot, which includes switching of the swing and support arm and the motor's axis of rotation. 
If the system experiences a disturbance such that it cannot reach the desired fixed point F or B, it will eventually come to rest in the Z configuration. Here, it can perform a ZB or ZF behavior to continue the forward brachiation via BF. 

\vspace{-0.2cm}
\subsection{Realization of Release \& Catch Behaviors}
\vspace{-0.1cm}
The passive gripper design was empirically optimized such that the gripper's interactions with the monkey bars can be achieved with a control heuristic on the elbow motor
\begin{Changed}
	, which depends on contact friction but is largely invariant to distance between the bars ($0.22-0.58$m).
\end{Changed}
The anti-clockwise rotation of the motor is referenced as positive as depicted in Figure~\ref{fig:2darm:coordinate}.


\subsubsection{Release} 
To simplify the control, BF and FB controllers are only engaged once the swing arm releases the bar.
For BR, a constant positive torque of 2.5 Nm is applied for at least 0.05 seconds. 
After this, if the elbow velocity surpasses 1.45 rad/s, the controller switches to BF brachiation. 
Empirical state data $\mathbf{x} = [\mathbf{q}, \mathbf{\dot{q}}]^T$ was collected over 20 trials at the point of controller transition. 
The state standard deviations $\bm{\sigma}^\text{BR}_{0} = [0.03, 0.03, 0.08, 0.11]^T$ were found to be relatively low, thus the trial mean values of the state $\mathbf{x}^\text{BR}_0 = [-0.63, -1.87, -0.63, 1.45]^T$ at this transition point were used as a reliable initial condition for controller generation for BF.
For FB swing, which starts with FR, 
a constant torque approach was insufficient due to the different contact angles of the hook on the bar. 
In order to clear the front bar, an initial high negative torque and a subsequent lower sustained positive torque is applied to lift the hook groove off the bar. 
Similar to BR, state data was collected at this transition point, analyzed, and used as the initial condition $\mathbf{x}^\text{FR}_0 = [0.51, 2.21, -0.63, 4.68]^T$ for the FB controllers with $\bm{\sigma}^\text{FR}_{0} = [0.03, 0.002, 0.42, 0.72]^T$. 


\subsubsection{Catch} 
The catch behavior is executed at the end of each atomic sub-behavior during 
continuous brachiation to provide a defined rotation point for the next brachiation. 
This is realized by applying a negative torque for 0.1~seconds with a magnitude of 0.8~Nm.
Duration and magnitude were chosen empirically such that the bar slides into the groove if the hook is slightly misplaced, but no movement is caused if the bar is already resting in the groove. 

\begin{figure}[tbp]
    \newcommand{\x}{0.8\linewidth}
    \centering
    \includegraphics[width=\x]{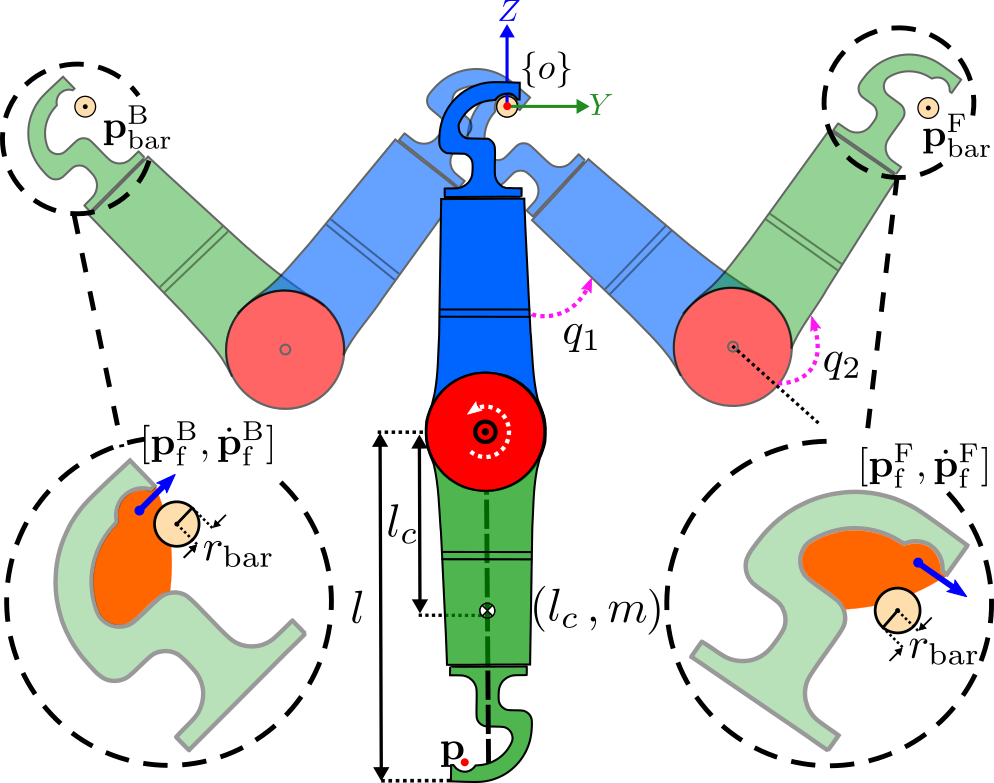}
    \caption{Schematic of the symmetric AcroMonk system with physical parameters $l=0.31, l_c=0.16$ m and $m=0.63$ kg}
    \label{fig:2darm:coordinate}
    \vspace{-0.5cm}
\end{figure}

\vspace{-0.3cm}
\subsection{Swing Behavior Generation}
\vspace{-0.1cm}
To complete the prerequisites for continuous brachiation, the atomic swing behaviors are generated 
using two different methods, namely trajectory optimization and RL.

\subsubsection{Trajectory Optimization}
\label{subsec:trajopt}
Finding the four atomic swing behaviors (ZF, ZB, FB, BF) for the AcroMonk system can be casted as a trajectory optimization problem:

\begin{subequations}
\begin{equation}
	\label{eq:gencostfunction}
	\begin{aligned}
		&\min_{\mathbf{x}, u} W T + \int_{0}^{T} \left(\mathbf{x}^T \mathbf{Q} \mathbf{x} + u^T R u\right) dt  \\
		&\mathrm{subject\ to}:
	\end{aligned}
\end{equation}
\vspace{-0.6cm}
\begin{gather}
	\dot{\mathbf{x}} = \mathbf{f}(\mathbf{x}, u) 	\label{eq:dynconstraint} \\
	|\mathbf{x}| \leq \mathbf{x}_\text{lim}, \ \ |u| \leq u_\text{lim}\label{eq:stateconstraint} \\
	\mathbf{x}(0) = \mathbf{x}_0, \ \ \ \mathbf{x}(T) = \mathbf{x}_f \label{eq:initfinstateconstraint} \\
	|| \mathbf{p} - \mathbf{p}^\text{F}_\text{bar} || > r_\text{bar}, \ \  || \mathbf{p} - \mathbf{p}^\text{B}_\text{bar} || > r_\text{bar}\label{eq:collisionconstraint}
\end{gather}
\end{subequations}
where the final cost term includes minimization of total trajectory time $T$ with weight $W$, and the running costs include 
a state regularization cost $\mathbf{x}^T \mathbf{Q} \mathbf{x}$ with $\mathbf{Q}=\mathbf{Q}^T \succeq 0$ and an 
effort regularization cost $u^T R u$ with $\mathbf{R}=\mathbf{R}^T \succ 0$.
The set of constraints include first order ODE (\ref{eq:dynconstraint}) form of system dynamics given by (\ref{eq:eom}), 
state and effort limits (\ref{eq:stateconstraint}),
initial and final values of the state (\ref{eq:initfinstateconstraint}), and
collision avoidance constraints (\ref{eq:collisionconstraint}) where $\mathbf{p}$ is the current position of the end-effector (EE) 
obtained via forward kinematics and $\mathbf{p}^\text{B}_\text{bar}, \mathbf{p}^\text{F}_\text{bar}, r_\text{bar}$ denote the space-fixed position of the left and right bars and their radii as shown in \autoref{fig:2darm:coordinate}. 
Direct Collocation~\cite{Betts2010} was used to find the optimal trajectories for the atomic behaviors using the Drake framework~\cite{tedrake2019drake}
with SNOPT~\cite{gill2005snopt} as the backend solver. 
The input trajectories are represented using a first-order hold trajectory while state trajectories are represented using a cubic spline interpolation. 

The hyperparameters of the running cost evaluated over $N=20$ knot points 
were empirically selected as $\mathbf{Q}=\text{diag}({0,0,1,1}), R = 100$ for all behaviors. The state and effort limits were conservatively chosen as $\mathbf{x}_\text{lim} = (2.09~\text{rad}, 2.88~\text{rad}, 10~\text{rad/s}, 10~\text{rad/s})^T$ and $u_\text{lim} = 3$ Nm to limit the search space of decision variables $(\mathbf{x}, u)$.  
The remaining hyperparameters are summarized in Table~\ref{tab:timing} for the four atomic behaviors.
The final state $\mathbf{x}_f$ for reaching the backward bar (valid for ZB and FB movements) is chosen via the (position and velocity level) inverse kinematics map such that the EE reaches the cartesian point $\mathbf{p}^B_f$ with velocity $\dot{\mathbf{p}}^B_f$, following which the passive dynamics of the system brings the bar into gripper's region of attraction (shown in orange in \autoref{fig:2darm:coordinate}) and settles the system to its stable fixed point. 
A similar argument holds for choosing $\mathbf{x}_f$ for reaching the front bar in case of ZF and BF movements.
It is crucial to minimize time in case of ZF and BF movements so that the EE reaches the point $\mathbf{p}^F_f$ (with velocity $\dot{\mathbf{p}}^F_f$) above the front bar with minimum number of swings. 

\begin{Changed}{
{
\begingroup
\vspace{-0.1cm}
\begin{table}[!htpb]
	\centering
	\setlength{\tabcolsep}{5pt}
	\renewcommand{\arraystretch}{0.0}
	\caption{Hyperparameters for trajectory optimization}
	\spaceskip=8pt
	\scriptsize
	\vspace{-0.1cm}
	\begin{tabular}{ccccc}
	\toprule
	\multicolumn{1}{c}{Behavior} & $\mathbf{x}_0$            & $\mathbf{x}_f$             & $W_t$ \\ \midrule
	ZB                           & $\mathbf{0}_{4 \times 1}$ & (-0.55, -1.97, -0.5, -3.0)& 0     \\ \midrule
	ZF                           & $\mathbf{0}_{4 \times 1}$ & (0.73, 1.92, -3.0, -2.5)   & 50    \\ \midrule
	FB     				         & $\mathbf{x}^\text{FR}_0$  & (-0.55, -1.97, -0.5, -3.0) & 0     \\ \midrule
	BF						     & $\mathbf{x}^\text{BR}_0$  & (0.73, 1.92, -3.0, -2.5)   & 50    & \\ \bottomrule
	\end{tabular}
	\label{tab:timing}
	\end{table}
\endgroup	
}
\vspace{-0.5cm}
}
\end{Changed}

\subsubsection{Reinforcement Learning}

\begin{figure}[ht]
	\centering
	\includegraphics[width=1.0\linewidth]{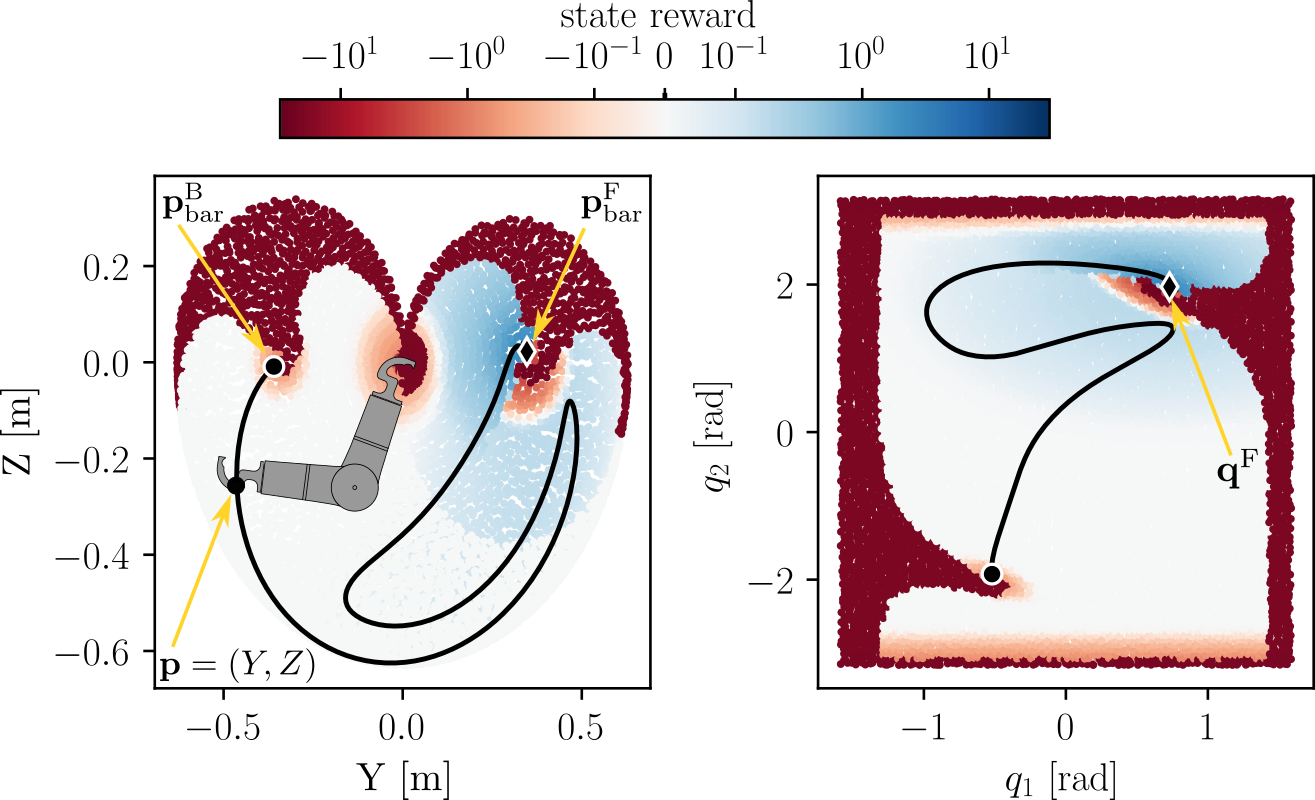}
	\caption{Reward visualization in task space (left) and joint space (right), excluding dynamic rewards $r^{-}_{u}$, $r^{-}_{\mathrm{vel}}$, and $r^{-}_{\dot{u}}$, and assuming $\dot{\mathbf{q}} = \mathbf{0}$. 
	Highly negative rewards correspond to configurations in collision. 
	The black line shows a trajectory generated by the torque controller after training.
	}
	\label{fig:reinf_learn_reward}
\end{figure}
\vspace{-0.2cm}

A BF controller was realized with model free RL, generating a policy $\pi$ which maps the observation $\mathbf{x}=[\mathbf{q}, \dot{\mathbf{q}}]^T$
to the torque $u$ directly applied to the motor, such that a 
reward function $r$ is maximized. The full reward $r$ is the sum of the terms detailed in Table \ref{tab:reward}.
To define reward terms in task space ($r^{-}_{c}, r^{-}_{sb}, r^{-}_{cb}, r^{-}_{tb}, r^{+}_{tb}$), the following functions are used 
\vspace{-0.2cm}
\begin{align}
	\begin{split}
		&g(\mathbf{p}, \mathbf{p}_\mathrm{bar}, d_{\mathrm{max}}) = H(d_{\mathrm{max}} - d) \left(\frac{d}{d_{\mathrm{max}}} - 1\right)^2 \\
		&h(\mathbf{p}, \mathbf{p_\mathrm{bar}}, d_{\mathrm{max}}, \mathbf{n}) = H(\langle d, \mathbf{n}\rangle) g(\mathbf{p}, \mathbf{p_\mathrm{bar}}, d_{\mathrm{max}})		
	\end{split}
\end{align}
where $H$ denotes the Heaviside function, $\mathbf{p}$ are coordinates of the swing arm end 
effector, $\langle\rangle$ denotes the scalar product, $\mathbf{n}$ defines a linear separatrix trough $\mathbf{p_\mathrm{bar}}$, $d = ||\mathbf{p} - \mathbf{p_\mathrm{bar}}||$, and $d_{\mathrm{max}}$ controls the region of influence of the term. 
In addition, reward terms are used in configuration space ($r^{+}_{tc}$), and dynamics penalties on the torque, velocity, and first derivative 
of effort ($r^{-}_{u}, r^{-}_{\mathrm{vel}}, r^{-}_{\mathrm{\dot{u}}}$) to generate controllers that can be safely executed on the hardware. 
Finally, reaching the target configuration $\mathbf{q}^{\text{F}}$ with an error smaller than $\Delta q = 0.05$ was rewarded ($r^{+}_{\mathrm{tar}}$).

	\begingroup
	\begin{table}[!htpb]
		\centering
		\setlength{\tabcolsep}{2pt}
		\renewcommand{\arraystretch}{1.1}
		\caption{Reward (+) and penalty (-) terms}
		\scriptsize
		\vspace{-0.1cm}
		\begin{tabular}{ll}
		\toprule
		Description          & Reward/Penalty term    \\ \midrule
		Collision term                                           & $r^{-}_{c} = -20$ if collision occurred    \\ \midrule
		Close approximation of back bar                          & $r^{-}_{sb} = g(\mathbf{p}, \mathbf{p}^\text{B}_\text{bar}, 10r_\text{bar})$    \\ \midrule
		Close approximation of support bar   		             & $r^{-}_{cb} = g(\mathbf{p}, [0, 0]^T, 17r_\text{bar})$     \\ \midrule
		Approaching the target bar from below   		         & $r^{-}_{tb} = -5h(\mathbf{p}, \mathbf{p}^\text{F}_\text{bar}, 15r_\text{bar}, [1.7, -1]^T)$     \\ \midrule
		\multirow{2}{*}{Approaching the target bar from above}
		  		         & $r^{+}_{tb} = \sum_kh(\mathbf{p}, \mathbf{p}^\text{F}_\text{bar}, d_k, [-1.7, 1]^T)$ \\
			& with $d = [30r_\text{bar}, 15r_\text{bar}, 10r_\text{bar}]$ \\ \midrule
		Configuration space   		                             & $r^{+}_{tc} = 0.2\exp(- 0.5\sum_i(q_i -  q_i^\text{F})^2)$     \\ \midrule
		Hardware torque limit 		                             & $r^{-}_{u} = -H(|u| - u_\text{lim}) (|u| - u_\text{lim})^2$     \\ \midrule
		Hardware velocity limit                                  & $r^{-}_{\mathrm{vel}} = -H(|\dot{q_2}| - 6) (|\dot{q_2}| - 6)^2$    \\ \midrule
		Smooth torque			                                 & $r^{-}_{\mathrm{\dot{u}}} = -0.001|u_t - u_{t-1}|$     \\ \midrule
		Reach final configuration       						 & $r^{+}_{\mathrm{tar}} = 30H(||\mathbf{q} - \mathbf{q}^\text{F}|| - \Delta q)$    \\ \bottomrule
		\end{tabular}
		\label{tab:reward}
		\vspace{-0.2cm}
		\end{table}
	\endgroup

Whereas calculating the reward requires information about the task space position $\mathbf{p}$ of the end effector, 
the observation of the policy only includes joint configurations and velocities. 
The reward function is visualized in Figure~\ref{fig:reinf_learn_reward}.

An episode was terminated if a collision occurred, or the maximal episode length of 2s or the target configuration with an error less than $\Delta q$ was reached. 
The system dynamics were simulated with MuJoCo~\cite{todorov2012mujoco} for training, at a simulation 
and control frequency of 250 Hz. 
Proximal Policy Optimization~\cite{schulman2017proximal} was used in the stable baselines~\cite{stable-baselines3} implementation with default parameters. 
To account for realistic measurement noise, normally distributed noise with $\sigma = 0.025$ was added to the state observations. Similar reward setups can be used to train controllers for all other atomic behaviors.


\vspace{\sectionspace}
\section{Behavior Control}
\label{sec:control}

Having generated optimal trajectories, these have to be 
tracked and stabilized during execution. 
In the case of RL, 
some deliberation is usually needed in tuning the simulation parameters 
for the policy
to perform well on the real system. The following section 
details the steps taken to realize the atomic swing 
behaviors and enable continuous and robust brachiation 
on the real robot.
\vspace{-0.5cm}
\subsection{Trajectory Tracking with PD}
\vspace{-0.1cm}
As a first method, we consider tracking the generated 
trajectories from \autoref{subsec:trajopt} with PD control for all 
atomic behaviors. 
The commanded torque from the state feedback for the actuated joint is computed using:
\vspace{-0.1cm}
\begin{equation}
\label{eq:pdcontrol}
\tau(t) = K_p (q_2(t) - q_2^*(t)) + K_d (\dot{q}_2(t) - \dot{q}_2^*(t)) + u^*(t)
\vspace{-0.1cm}
\end{equation}

Here, ${}^*$ denotes the nominal trajectories. 
We chose the controller gains $K_p = 100$ and $K_d = 2$ empirically at a control frequency of 300 Hz.
An idling time of 0.1 seconds was used before engaging any Catch behavior to leave enough time for the 
catching hook to make contact with the target bar.

\vspace{-0.4cm}

\subsection{Trajectory Tracking with TVLQR}
\vspace{-0.1cm}
As an alternative,  Time-Varying Linear Quadratic Regulator (TVLQR)~\cite{underactuated} control was also used to stabilize the nominal trajectories. 
TVLQR aims to minimize the error coordinates $\mathbf{\bar{x}} = (\mathbf{x} - \mathbf{x}^*)$ and $\mathbf{\bar{u}} = (\mathbf{u} - \mathbf{u}^*)$ 
, where ${}^*$ denote states of the nominal trajectory. 
For this, a time-varying linearization using a Taylor series approximation is performed, resulting in a time-varying linear system in the error coordinates:
\vspace{-0.1cm}
\begin{equation}
\label{eq:timevarylinearsys}
\mathbf{\dot{x}} = \mathbf{A}(t) \mathbf{\bar{x}}(t) - \mathbf{B}(t) \mathbf{\bar{u}}(t)
\vspace{-0.1cm}
\end{equation}
The quadratic cost function is defined as:
\vspace{-0.1cm}
\begin{equation*}
J = \mathbf{\bar{x}}^T(t) \mathbf{Q}_f \mathbf{\bar{x}}(t) + \int_{0}^{t_f} \left( \mathbf{\bar{x}}^T(t) \mathbf{Q} \mathbf{\bar{x}}(t) + \mathbf{\bar{u}}^T(t) \mathbf{R} \mathbf{\bar{u}}(t)\right) \,\, dt
\vspace{-0.1cm}
\end{equation*}
where $\mathbf{Q}=\mathbf{Q}^T \succeq 0$, $\mathbf{Q}_f=\mathbf{Q}_f^T \succeq 0$ and $\mathbf{R}=\mathbf{R}^T \succ 0$. 
The optimal cost-to-go can be written as a time-varying quadratic term and the controller gain $\mathbf{K}(t)$ be found by solving the differential Riccati Equation \cite{bertsekas2012dynamic}. 
The final control law is then of the form:
\vspace{-0.2cm}
\begin{equation}
\tau(t) = \mathbf{u}(t) = \mathbf{u}^*(t) - \mathbf{K}(t) (\mathbf{x} - \mathbf{x}^*).
\vspace{-0.2cm}
\end{equation}
The hyperparameters for the TVLQR-stabilized BF behavior were empirically selected as $Q=[0.01, 5, 0.01, 0.1]$, $Q_f=[0.04, 20, 0.04, 0.4]$, and $R = 5$.
These parameters worked for both swing arms. 
TVLQR stabilization was run at 260 Hz, slightly slower than PD, due to the 
extra computational step to find the closest point of the current state to 
the target trajectory. For continuous brachiation, in addition to the 0.1 
second idling time before each catch, an additional 0.1 s pause between 
successive BF behaviors was introduced, since the method was more susceptible 
to deviations in the initial condition after BR. All other atomic behaviors 
can be stabilized by TVLQR, but we focus here on BF without loss of generality.

\vspace{-0.5cm}
\subsection{Model Free RL Control}
\vspace{-0.1cm}
In contrast to the previous methods, RL trains the mapping of observations to torque control directly in simulation, not following a precomputed target trajectory. 
For direct torque control, there is a high demand on simulation accuracy for 
successful simulation to reality transfer.
To ensure realistic damping losses, trajectories of $\mathbf{q}, \dot{\mathbf{q}}, \tau$ from a BF swing via trajectory tracking with PD were recorded. Simulated trajectories $\mathbf{q}_\text{sim}, \dot{\mathbf{q}}_\text{sim}$ were obtained by replaying the recorded torques in simulation. 
The damping parameters of the support hook contact on the bar and the motor were optimized such that the deviations $\mathbf{q} - \mathbf{q}_\text{sim}, \dot{\mathbf{q}} - \dot{\mathbf{q}}_\text{sim}$ are minimized, 
following~\cite{kaspar2020sim2real}. The SHGO global optimizer (SciPy) yielded damping values of $\approx 0.044$ for the hook contact and $\approx 0.06$ for the motor. Furthermore, the BF controller was trained only for the swing arm connected to the motor housing. 
For BF with the other arm, the torque commands were scaled empirically by a factor of 0.92. The trained policy network was converted to a numpy function for deployment on the on-board computer. The controller 
was run at 80 Hz on the real system. Although capable to run faster, higher frequencies made the policy less stable, probably due to 
a higher impact of sensor noise. For continuous 
brachiation, the 
idling time before Catch was set to 0.2 s and the pause between subsequent BF 
behaviors to 0.5 s, both to the same end of giving the system enough time to 
settle and ensure low deviations from the expected initial condition after hook release.

\vspace{\sectionspace}
\section{Results \& Discussion}
\label{sec:results}

{    \begingroup
    \begin{table*}[htbp]
        \setlength{\tabcolsep}{0pt}
        \renewcommand{\arraystretch}{0.3}	
        \newcommand{\xscale}{0.15\linewidth}
        \centering
        \begin{tabular}{cccccc}
             \includegraphics[width=\xscale]{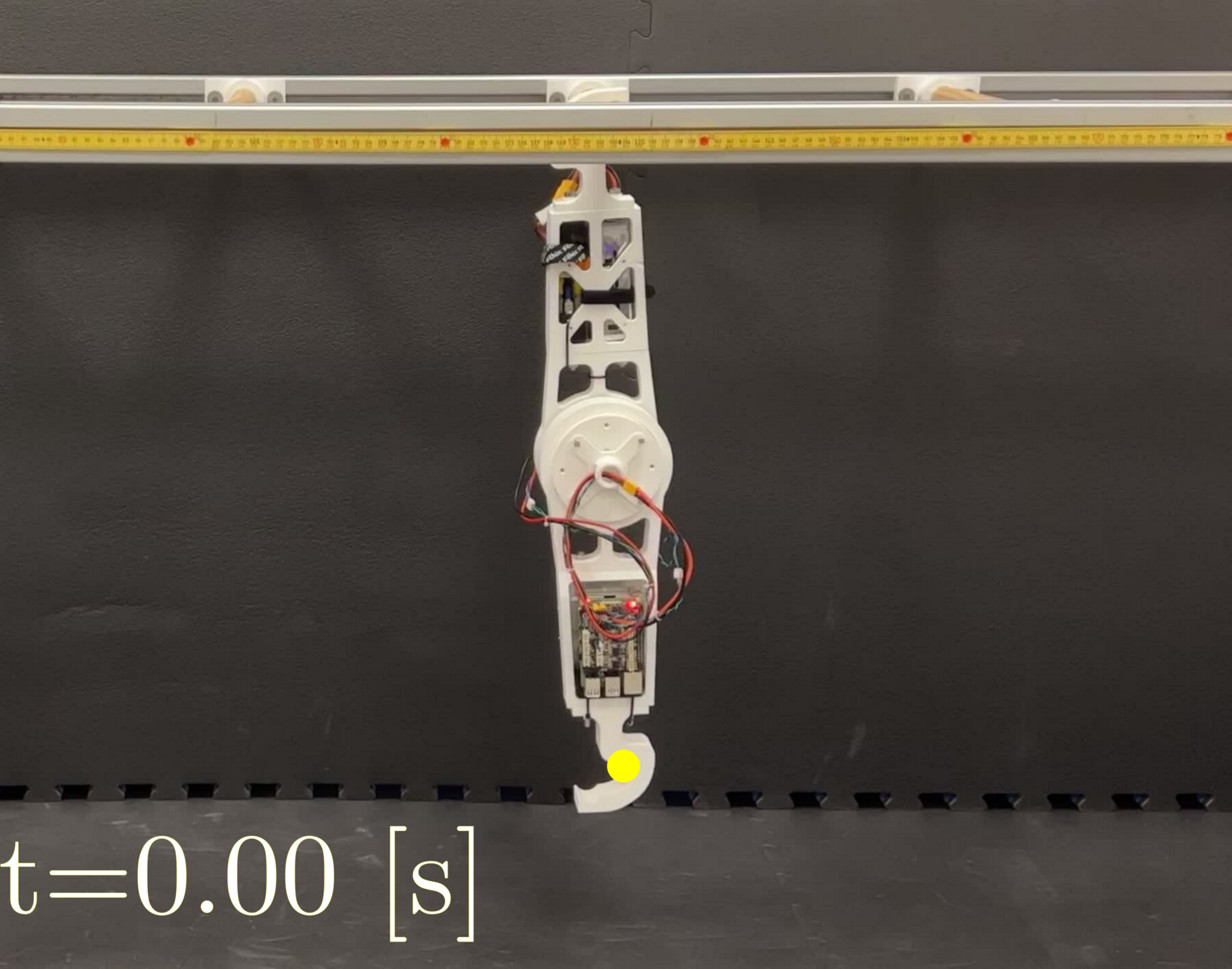}&
             \includegraphics[width=\xscale]{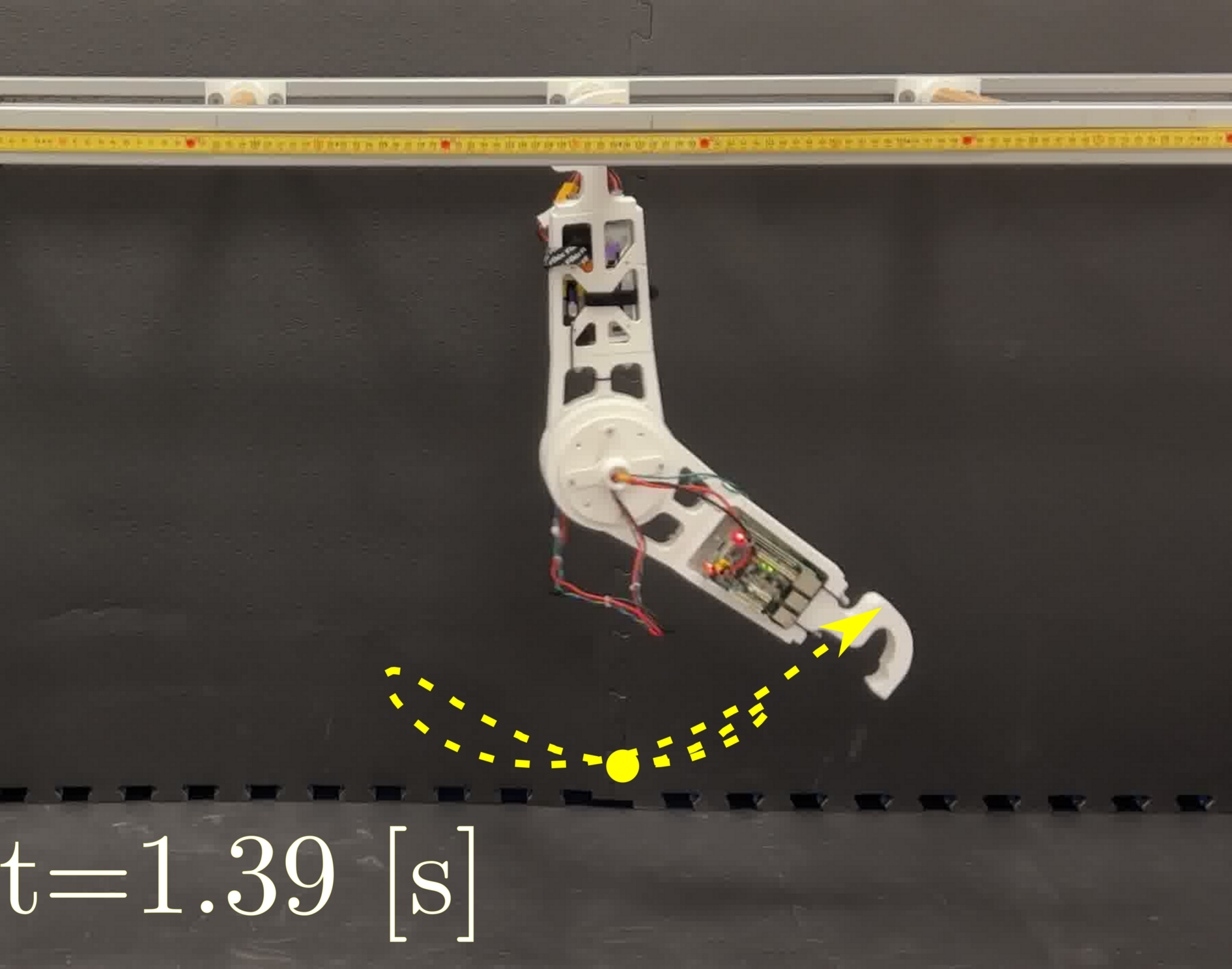}&
             \includegraphics[width=\xscale]{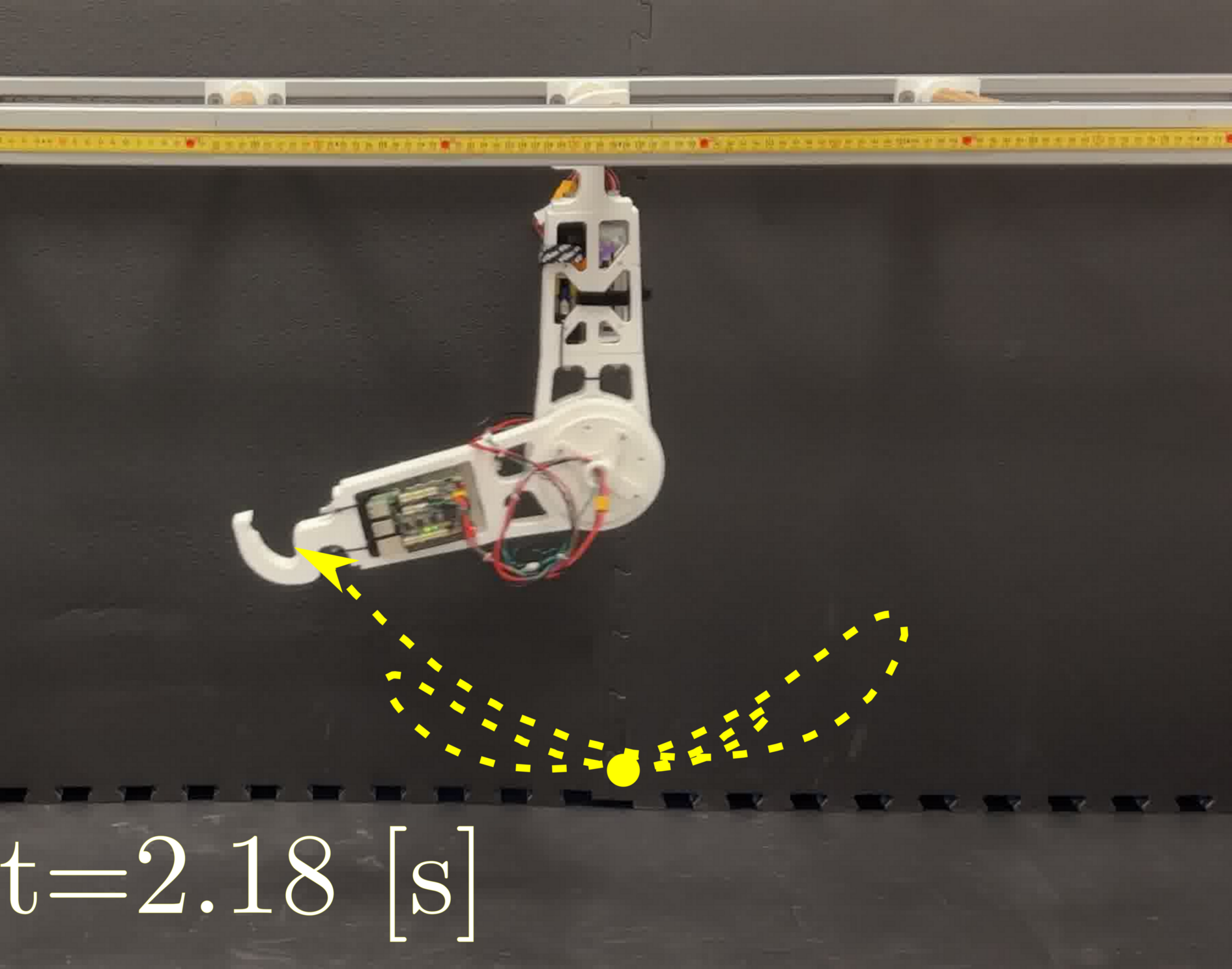}&
             \includegraphics[width=\xscale]{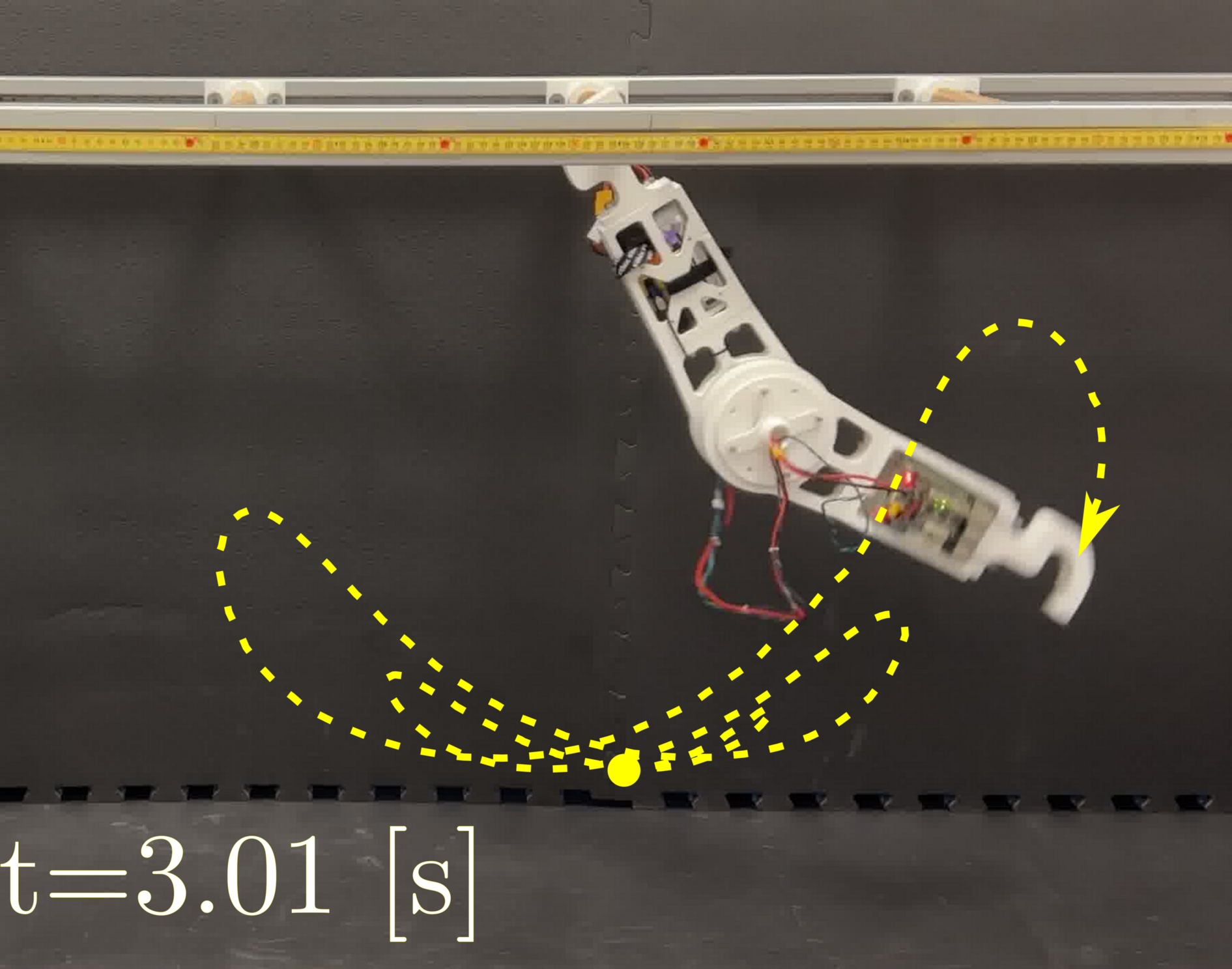}&
             \includegraphics[width=\xscale]{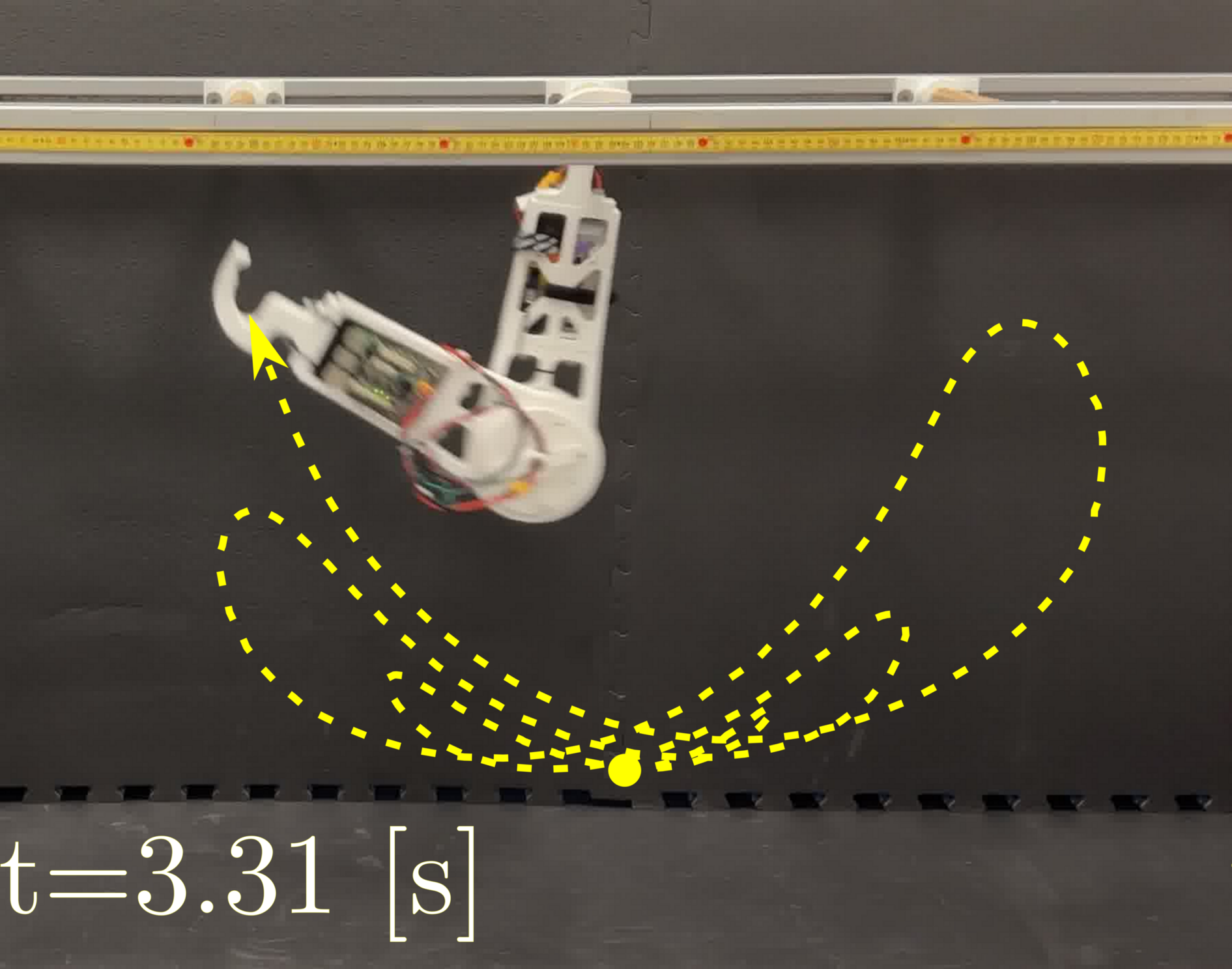}& 
             \includegraphics[width=\xscale]{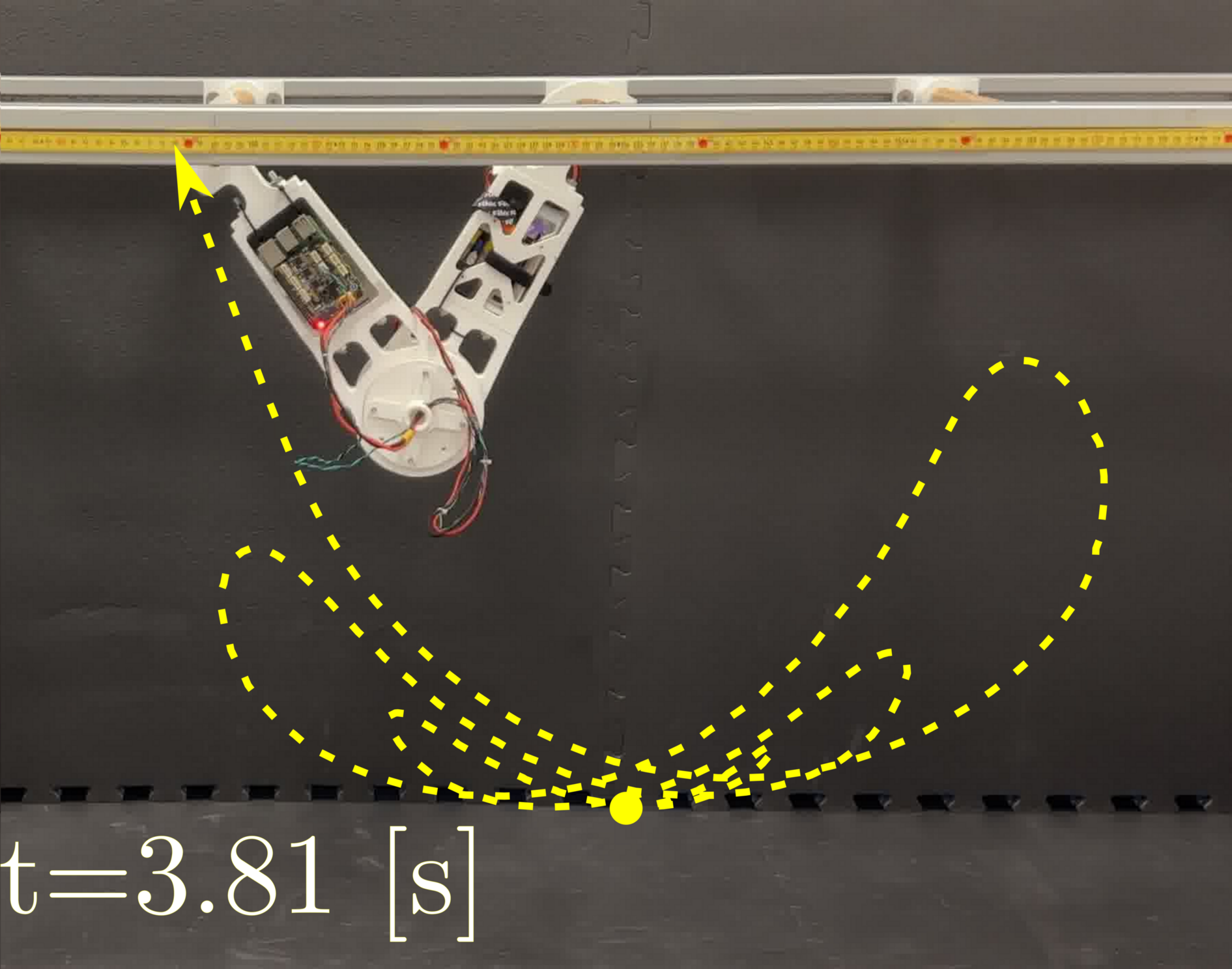} 		 
              \\
             \includegraphics[width=\xscale]{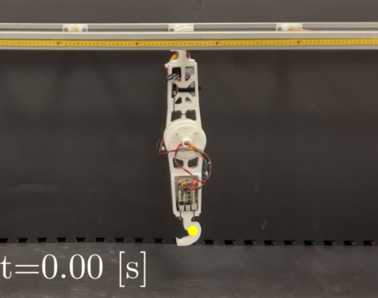}&
             \includegraphics[width=\xscale]{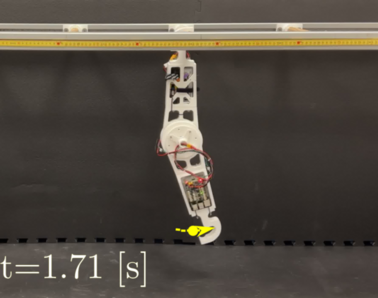}&
             \includegraphics[width=\xscale]{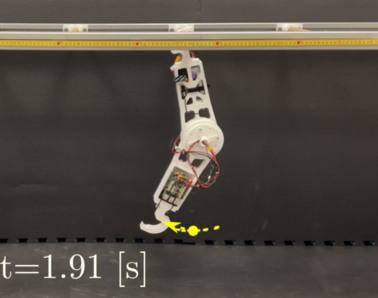}&
             \includegraphics[width=\xscale]{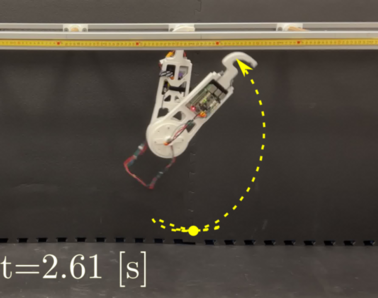}&
             \includegraphics[width=\xscale]{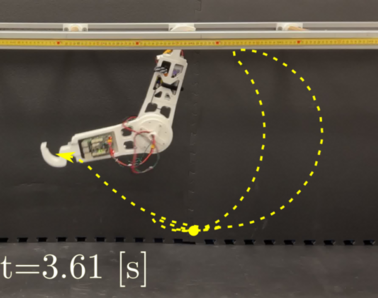}&
             \includegraphics[width=\xscale]{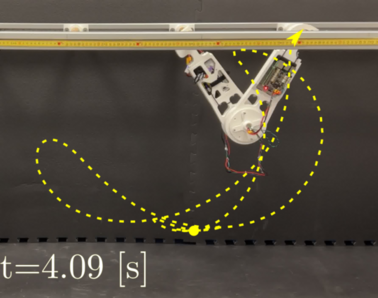}
             \\
             \includegraphics[width=\xscale]{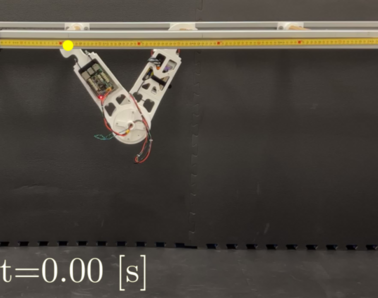}&
             \includegraphics[width=\xscale]{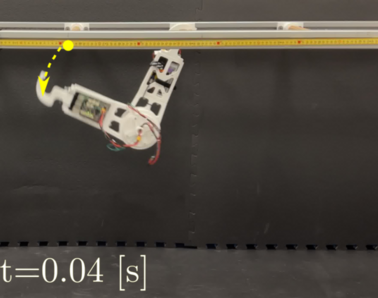}&
             \includegraphics[width=\xscale]{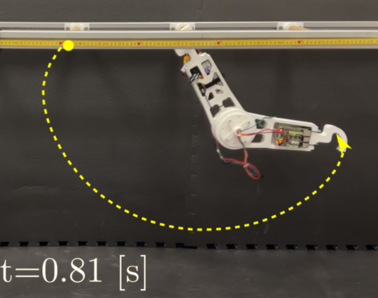}&
             \includegraphics[width=\xscale]{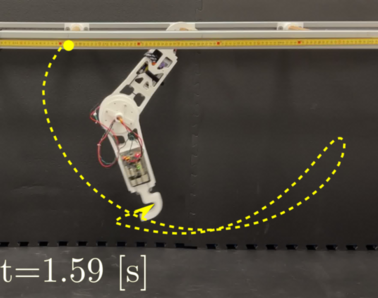}&
             \includegraphics[width=\xscale]{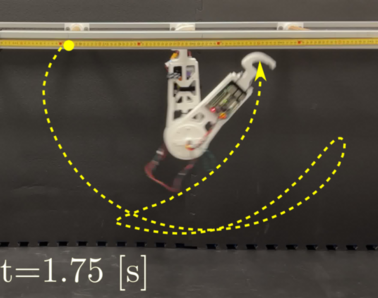}&
             \includegraphics[width=\xscale]{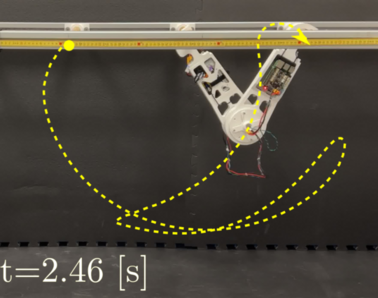}
             \\
             \includegraphics[width=\xscale]{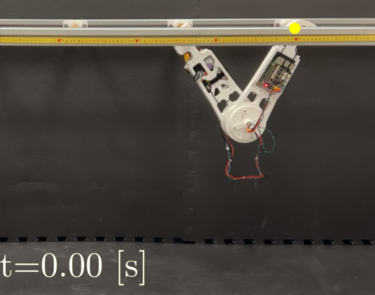}&
             \includegraphics[width=\xscale]{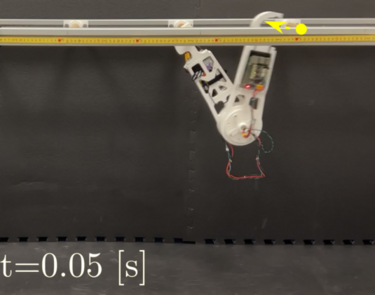}&
             \includegraphics[width=\xscale]{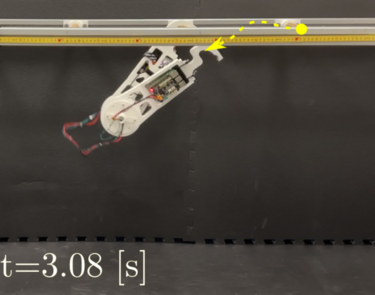}&
             \includegraphics[width=\xscale]{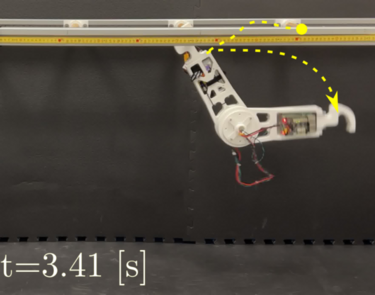}&
             \includegraphics[width=\xscale]{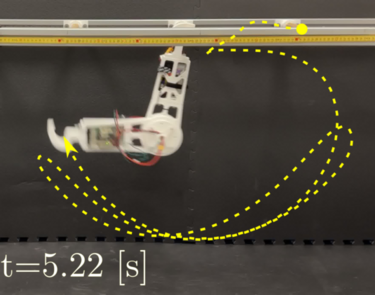}&
             \includegraphics[width=\xscale]{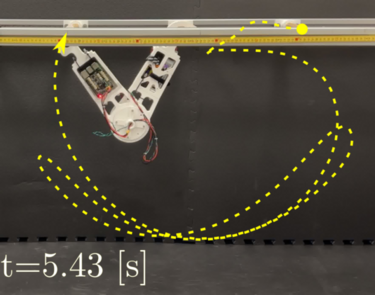}		 		  	          
        \end{tabular}
        \captionof{figure}{Snapshots of executions of the atomic behaviors ZB~(top), ZF~(second row), BF~(third row), and FB~(bottom)}
        \label{fig:atomic_snapshots}
    \end{table*}
    \endgroup
    }

The four atomic behaviors BF, FB, ZB, and ZF were realized on the hardware demonstrator with Trajectory optimization (Traj Opt) + PD. 
Figure~\ref{fig:atomic_snapshots} shows snapshots of the successful execution of these behaviors. This allows for continuous, 
bidirectional brachiation including recovery from disturbances, as 
shown in the supplementary video.

Whereas in principle all behaviors can be achieved by different 
control methods (see Section \ref{sec:trajopt}), we use the example of BF 
to benchmark the performance 
of Traj Opt + PD, Traj Opt + TVLQR, and RL policy-based control. All methods achieved high repeatability of the single BF behavior with a 100\% success rate over five different trials. 
To simulate instantaneous disturbances due to collisions with the environment, a cardboard box (with dimensions $13 \times 8 \times 28$ cm and weight $160$ g) was placed on the ground in the swing path of the arm, roughly below the target bar. 
Here, only Traj Opt + PD could reliably recover with a success rate of 100\%. 
Traj Opt + TVLQR recovered in 4/5 cases and the RL controller in 1/5 cases. 
Robustness to mass uncertainty was assessed by attaching a 200 g weight to the swing arm. 
The Traj Opt + PD controller compensated for this mismatch reliably, whereas both Traj Opt + TVLQR and RL failed in 5/5 tests. 

To assess continuous brachiation performance, we benchmark timing and energy expenditure of five consecutive forward brachiations with all three control methods. 
Figure~\ref{fig:5x_brach} shows the full maneuver's positions, velocity, and torque trajectories. 
Table~\ref{tab:perfcontrol} summarizes benchmark values for maximum torque usage, trajectory tracking performance, overall energy consumption, and duration. 
The root-mean-square error of trajectory tracking is low for TVLQR and PD. 
This metric does not apply to RL since the policy does not track
a trajectory. 
The peak torque is the lowest for TVLQR and highest for RL.  
The lowest energy consumption was achieved by RL for the whole maneuver, whereas PD had a considerably higher energy demand. 
The total time of transport is highest for RL, even though the 
controller needs the shortest time to complete one swing. 
The reason lies in the RL-based controller's sensitivity to disturbances and uncertainties during the maneuver. 
Therefore, longer pauses in comparison to the other methods between each brachiation maneuver are introduced to let the system settle down. 


\begingroup
\begin{table*}[htbp]
	\setlength{\tabcolsep}{0pt}
	\renewcommand{\arraystretch}{0.0}	
	\newcommand{\xscale}{0.33\linewidth}
     \newcommand{\overlap}{-0.495cm}
    \centering
    \begin{tabular}{cccccc}
         \vspace{\overlap}
         \includegraphics[width=\xscale]{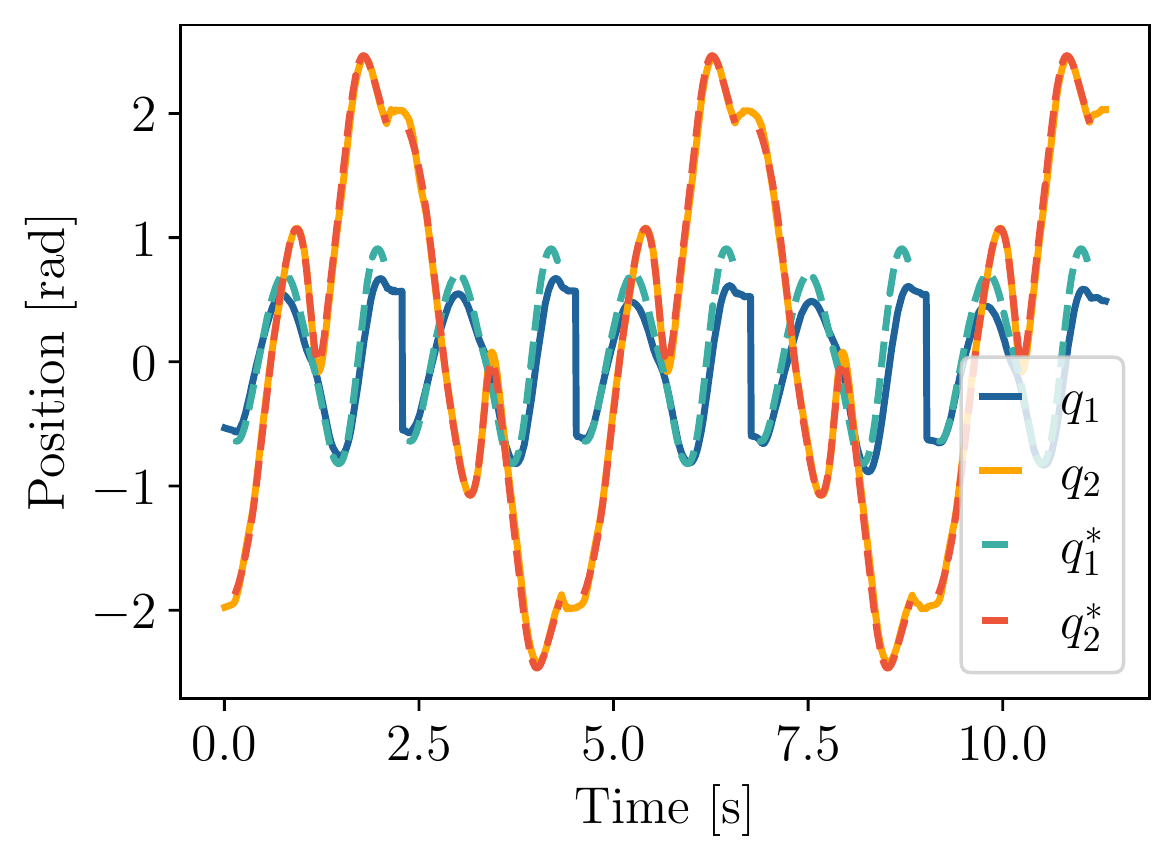}&
         \includegraphics[width=\xscale]{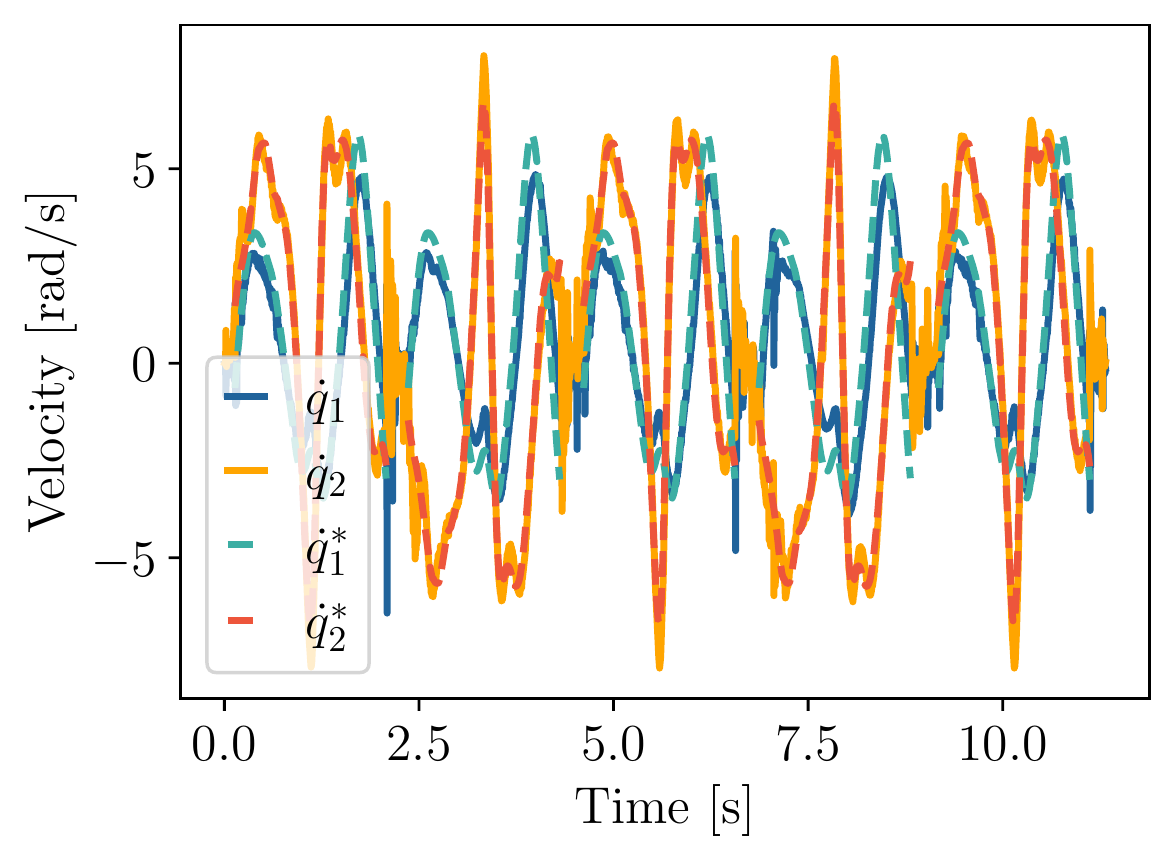}&
         \includegraphics[width=\xscale]{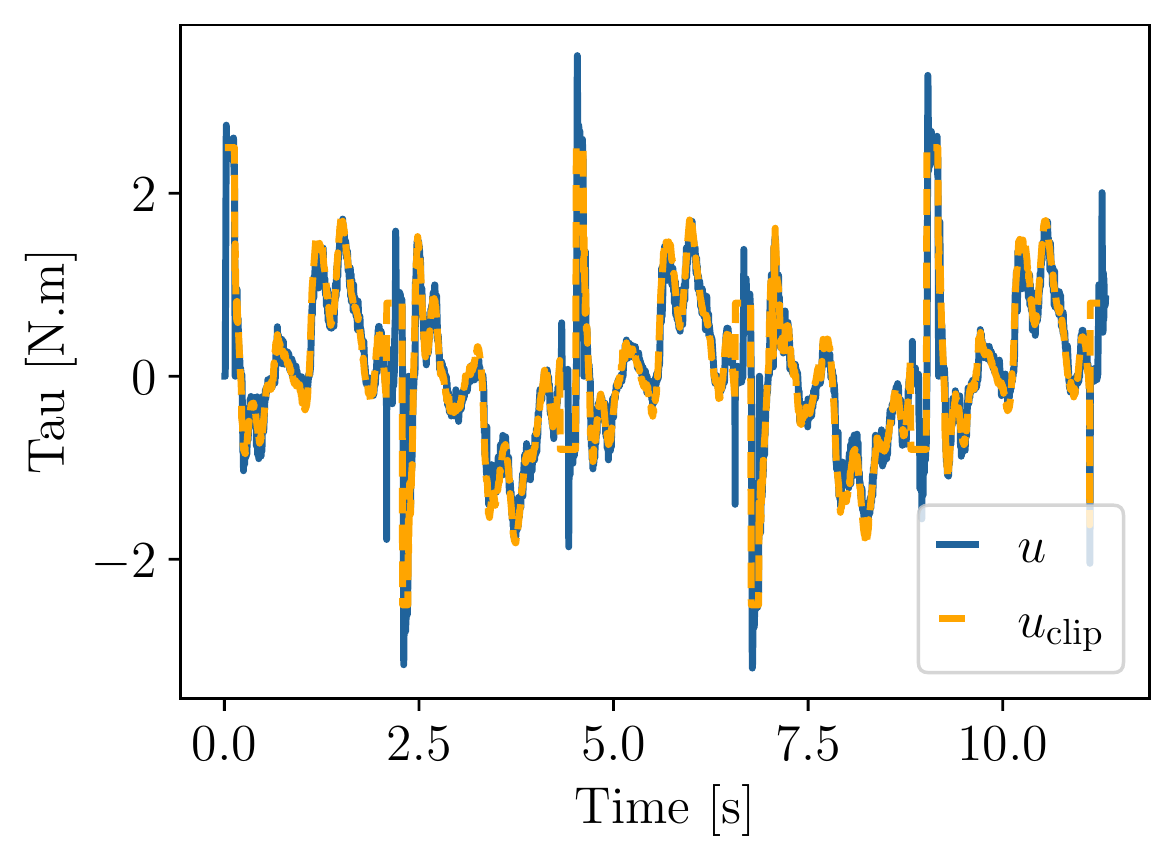}		 
          \\ \vspace{\overlap}
		  \includegraphics[width=\xscale]{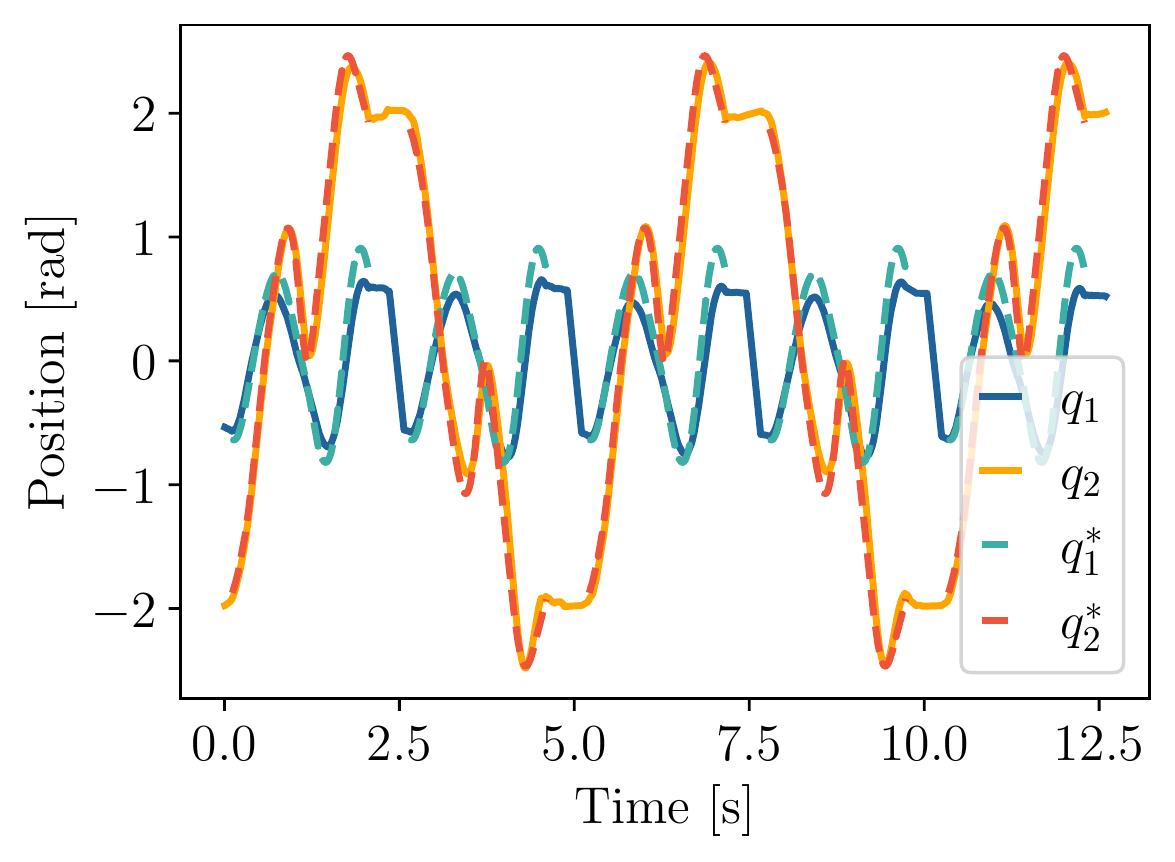}&
		  \includegraphics[width=0.34\linewidth]{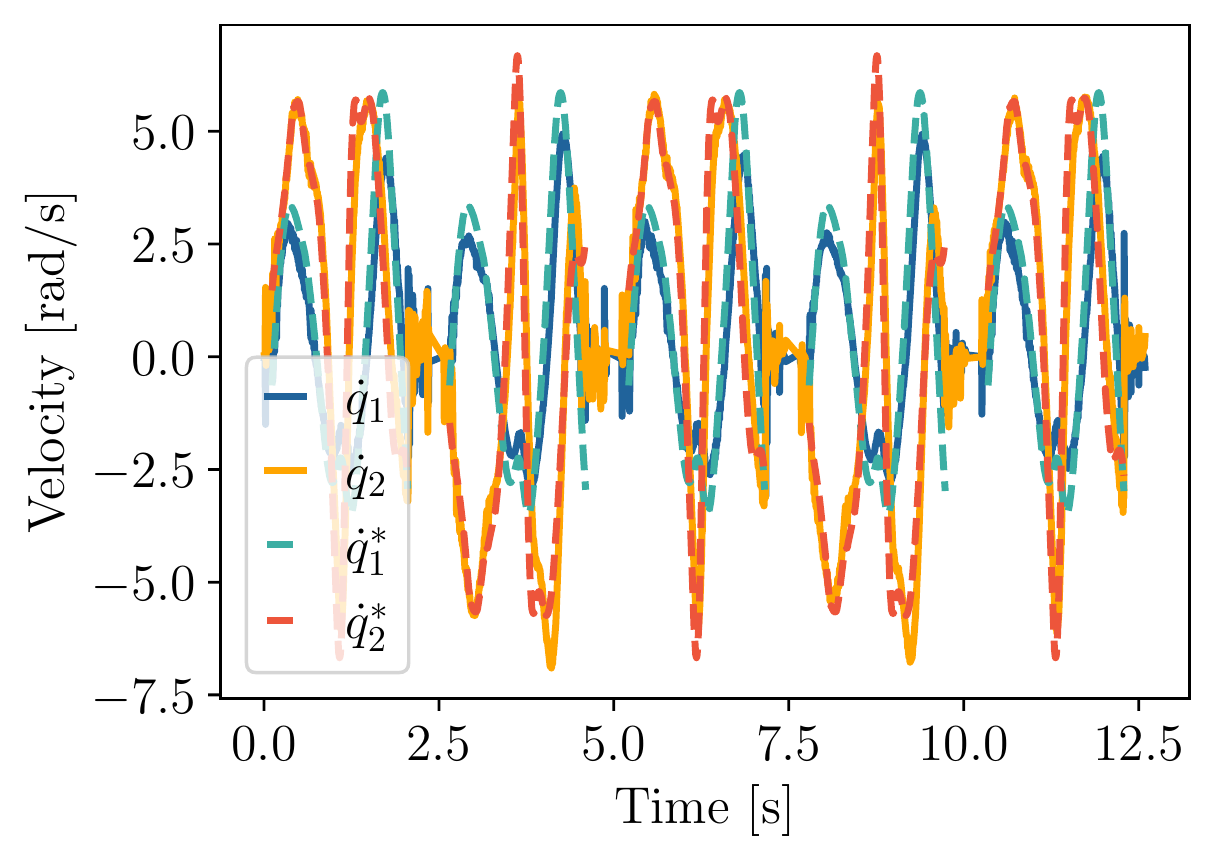}&
		  \includegraphics[width=\xscale]{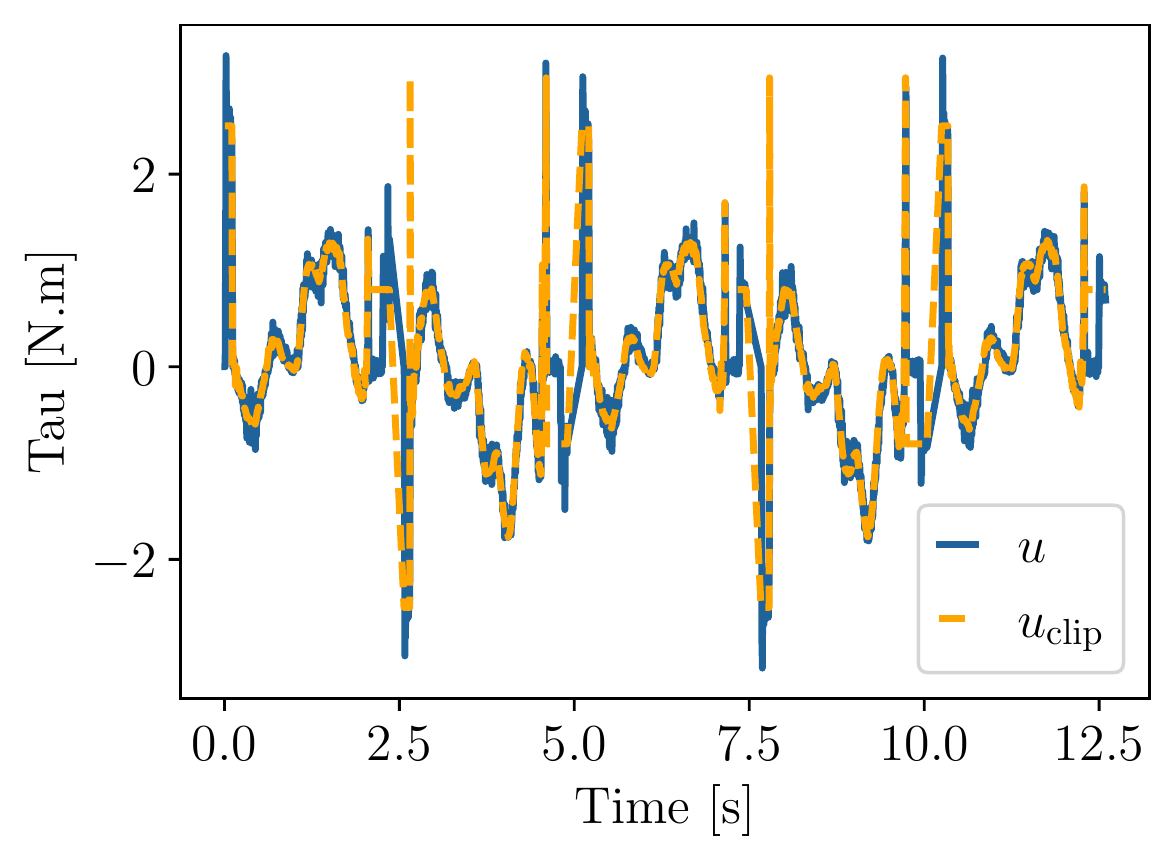}
          \\ 
         \includegraphics[width=\xscale]{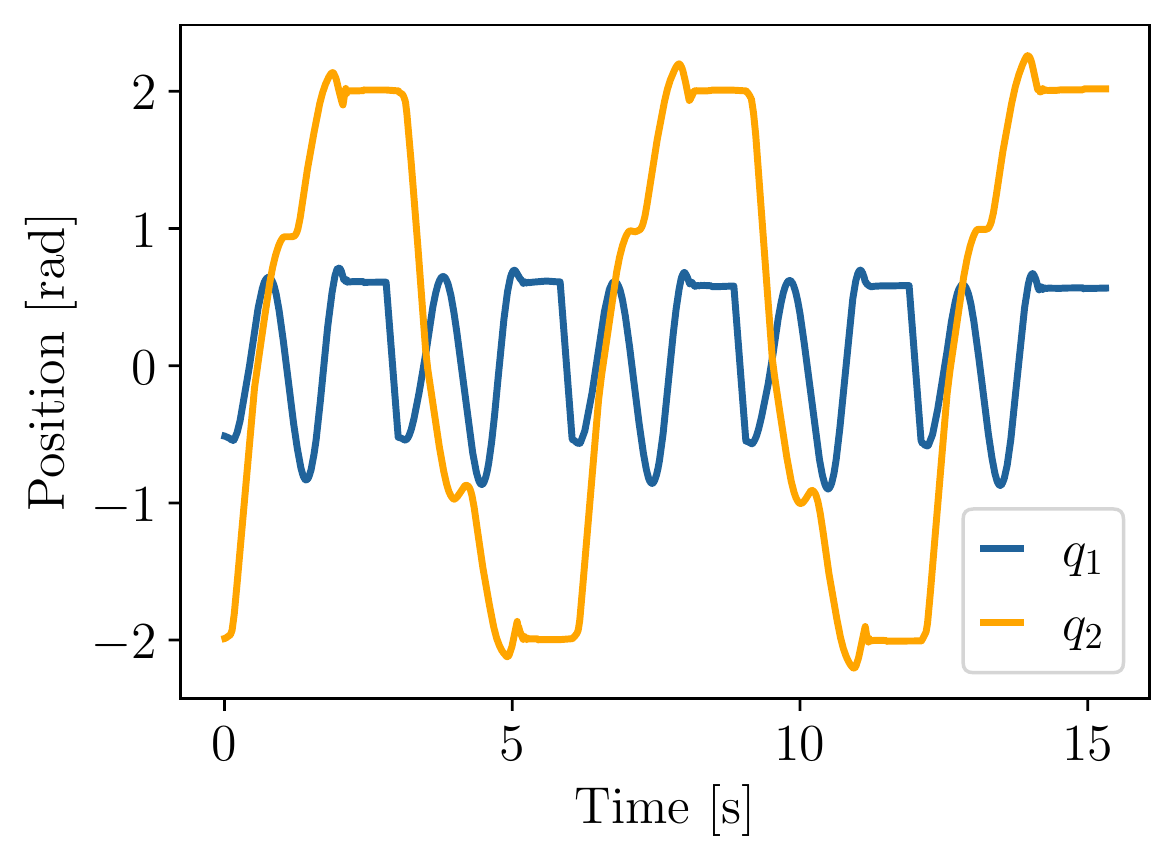}&
         \includegraphics[width=0.34\linewidth]{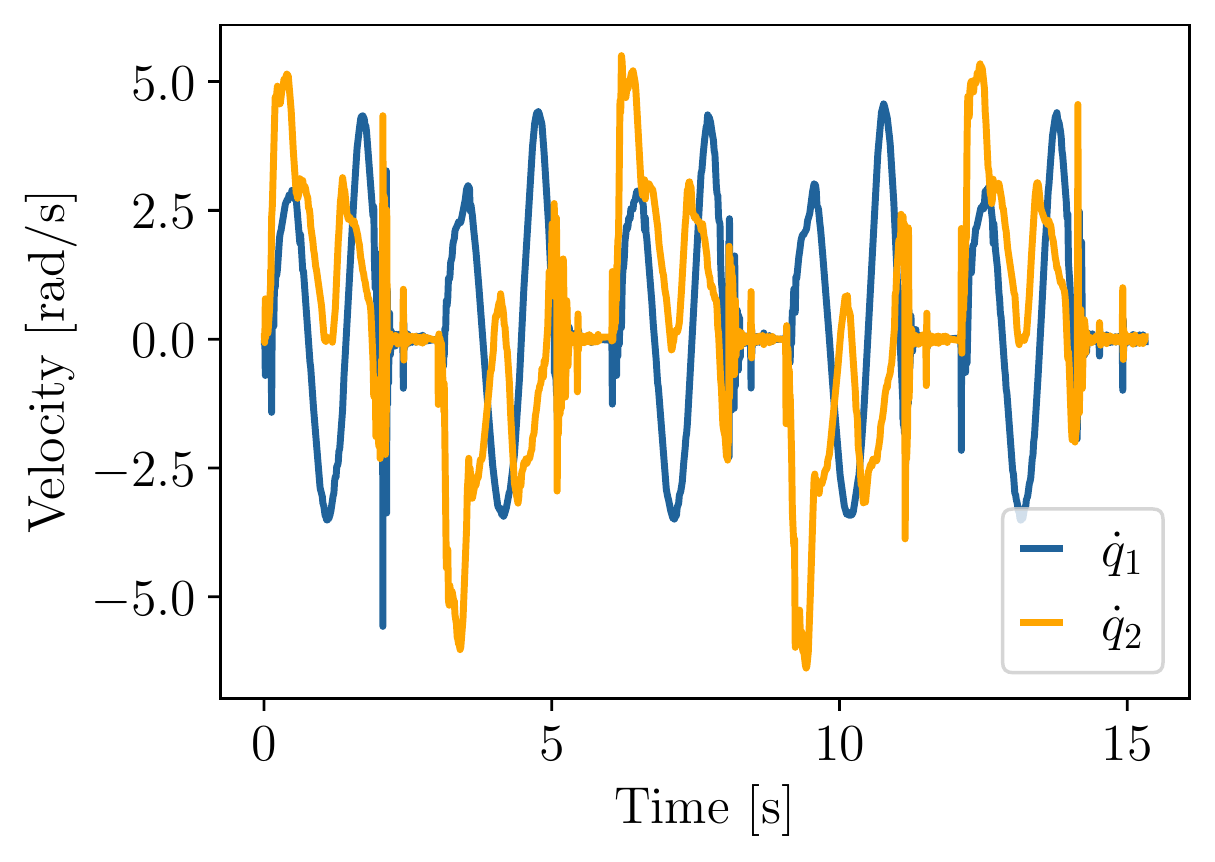}&
         \includegraphics[width=\xscale]{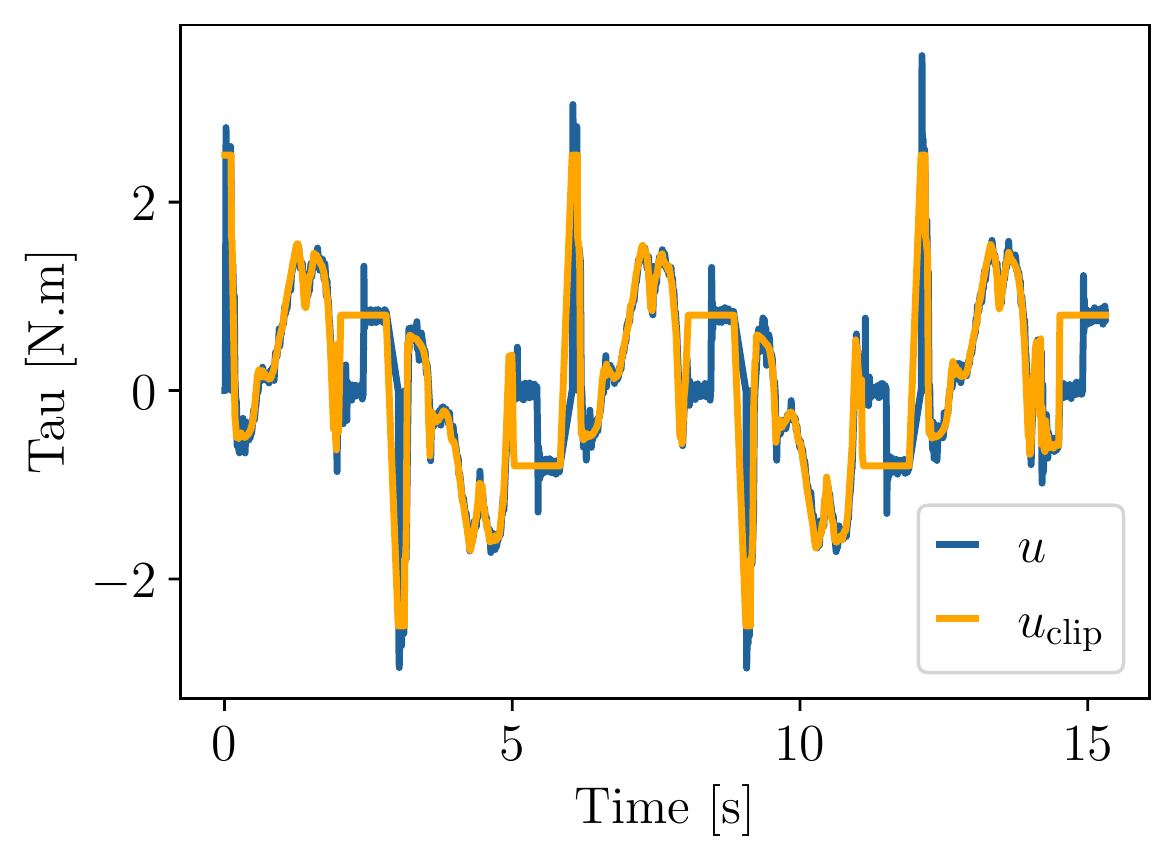}
    \end{tabular}
    \captionof{figure}{Recorded data of joint positions, velocities, and torque for five consecutive BF behaviors, i.e. six bars, with $0.34\ m$ gap.	Up: PD control, middle: TVLQR, bottom: RL control. 
	For PD and TVLQR, the desired trajectories are also shown.}

    \label{fig:5x_brach}
    \vspace{-0.5cm}
\end{table*}
\vspace{-0.1cm}
\begin{table}[!htpb]
\scriptsize
\caption{Performance characteristics of controllers}
\vspace{-0.1cm}
\begin{tabular}{ccccc}
	\toprule	
	\multicolumn{2}{c}{Metrics / Controller}                  & PD     & TVLQR   & RL     \\ \midrule
	Max. Abs. Torque (Nm)                   &                & 3.506   & 3.231   & 3.55   \\ \midrule
	\multirow{3}{*}{Average RMS Error}      & Pos (rad)      & 0.119   & 0.163   & -      \\
											& Vel (rad/s)    & 0.682   & 0.952   & -      \\
											& Torque (Nm)    & 0.709   & 0.430   & -      \\ \midrule	
	\multirow{2}{*}{Total Energy (Joule)}   & 5x Brachiation    & 8.939   & 8.640   & 6.995  \\
											& Single swing   & 1.645   & 1.556   & 1.167  \\ \midrule
	\multirow{2}{*}{Transport Duration (s)} & 5x Brachiation    & 11.322  & 12.592  & 15.310 \\
											& Single swing   & 1.944   & 1.944   & 1.894  \\ \midrule
    \multirow{2}{*}{Success Rate ($\%$)} & 5x Brachiation    & 100  & 100  & 10 \\
    & Single swing   & 100   & 100   & 100  \\
\bottomrule
\end{tabular}
\label{tab:perfcontrol}
\vspace{-0.5cm}
\end{table}

The results show that AcroMonk is an easily controllable system, despite its underactuation and passive gripper design. Realizing 
all atomic dynamics behaviors leads to successful and robust 
brachiation, while the system can recover from disturbances. 
The relative ease of controllability can be attributed to the balanced design and the novel, grooved grippers, that provide a well-determined starting point for each behavior. 

The successful performance of five forward brachiation motions with all three control strategies is a novelty in the literature for a system with passive grippers and only one motor.
The various controller types in this experiment showcase the advantages and disadvantages of different 
state of the art strategies.
Simple PD trajectory stabilization performed well and indeed proved to be most robust to external disturbances.
Given the design choices, this is not surprising since PD control will always force the trajectory back on track, provided enough torque is available.
TVLQR incorporates a model of the system to track the desired trajectory, which, if it does not match the actual setup, e.g., when an unknown mass is added, will lead to sub-optimal performance. On the other hand, it can be more energy efficient due to incorporation of model knowledge. 
Combining an optimized simulation model with RL resulted in the most energy efficient controller, which 
is however also most susceptible to deviations from trained states, resulting 
in longer necessary pauses between behaviors and poor generalization to disturbances or long traversals. 
\vspace{-0.05cm}

\vspace{\sectionspace}
\section{Conclusion}
\label{sec:conclusion}
\vspace{-0.07cm}
With AcroMonk, we present a novel canonical underactuated system for studying brachiation. 
Due to the grooved gripper design, it is easily and reliably controllable, making it the first system of such a low complexity to achieve \begin{Changed}{multiple}\end{Changed} consecutive brachiation motions. 
The readily available components and straightforward assembly make it a suitable reference system for underactuated robotics research. 
Our future work will focus on \begin{Changed}{the following}\end{Changed} issues.
Despite some success, we were not yet able to produce reliable backward brachiation. 
The release behavior in this configuration is much harder to perform since it requires lifting the swing arm hook up from the bar leading to a mean initial condition of front release with higher standard deviation $(\bm{\sigma}^\text{FR}_{0} > \bm{\sigma}^\text{BR}_{0})$. 
Also, due to the single-motor design, the desired support arm may seldomly unhook instead during this maneuver. 
To solve this problem, we are working on an improved gripper design with beveled edges to reduce the force required for unhooking. 
We are also working on realizing even more dynamic behavior such as automatic release during continuous brachiation and ricocheting exploiting impacts during the kinodynamic planning. We already observed that a well adjusted impact force on the target bar can directly unhook the support arm, resulting in even smoother and more dynamic brachiation. 
Considering ricocheting, we could also generate brachiation in a single swing with a short flight phase when removing the torque limits of the controller. 
While this was not yet safely reproducible, it shows that the system is in principle capable of such behavior. 
\begin{Changed}
    Finally, brachiation over irregularly placed bars is another challenge to be tackled in future. 
\end{Changed}
The design and controllers discussed in this paper have been open-sourced (\href{https://github.com/dfki-ric-underactuated-lab/acromonk}{https://github.com/dfki-ric-underactuated-lab/acromonk}) to support education and research of brachiation with easy to implement hardware.

\vspace{\sectionspace}
\vspace{-0.05cm}
\bibliographystyle{IEEEtran}
\begin{samepage}
\bibliography{references}

\begin{thebibliography}{10}
\providecommand{\url}[1]{#1}
\csname url@rmstyle\endcsname
\providecommand{\newblock}{\relax}
\providecommand{\bibinfo}[2]{#2}
\providecommand\BIBentrySTDinterwordspacing{\spaceskip=0pt\relax}
\providecommand\BIBentryALTinterwordstretchfactor{4}
\providecommand\BIBentryALTinterwordspacing{\spaceskip=\fontdimen2\font plus
\BIBentryALTinterwordstretchfactor\fontdimen3\font minus
  \fontdimen4\font\relax}
\providecommand\BIBforeignlanguage[2]{{%
\expandafter\ifx\csname l@#1\endcsname\relax
\typeout{** WARNING: IEEEtran.bst: No hyphenation pattern has been}%
\typeout{** loaded for the language `#1'. Using the pattern for}%
\typeout{** the default language instead.}%
\else
\language=\csname l@#1\endcsname
\fi
#2}}

\bibitem{fukuda-first}
T.~Fukuda, H.~Hosokai, and Y.~Kondo, ``Brachiation type of mobile robot,'' in
  \emph{Fifth International Conference on Advanced Robotics' Robots in
  Unstructured Environments}.\hskip 1em plus 0.5em minus 0.4em\relax IEEE,
  1991, pp. 915--920.

\bibitem{fukuda-swing}
F.~Saito, T.~Fukuda, and F.~Arai, ``Swing and locomotion control for a two-link
  brachiation robot,'' \emph{IEEE Control Systems Magazine}, vol.~14, no.~1,
  pp. 5--12, 1994.

\bibitem{meghdari2013minimum}
A.~Meghdari, S.~M.~H. Lavasani, M.~Norouzi, and M.~S.~R. Mousavi, ``Minimum
  control effort trajectory planning and tracking of the cedra brachiation
  robot,'' \emph{Robotica}, vol.~31, no.~7, pp. 1119--1129, 2013.

\bibitem{davies2018tarzan}
E.~Davies, A.~Garlow, S.~Farzan, J.~Rogers, and A.-P. Hu, ``Tarzan: Design,
  prototyping, and testing of a wire-borne brachiating robot,'' in \emph{2018
  IEEE/RSJ International Conference on Intelligent Robots and Systems
  (IROS)}.\hskip 1em plus 0.5em minus 0.4em\relax IEEE, 2018, pp. 7609--7614.

\bibitem{farzan-tvlqr}
S.~Farzan, A.~P. Hu, E.~Davies, and J.~Rogers, ``{Feedback motion planning and
  control of brachiating robots traversing flexible cables},''
  \emph{Proceedings of the American Control Conference}, vol. 2019-July, pp.
  1323--1329, 2019.

\bibitem{yang2019design}
S.~Yang, Z.~Gu, R.~Ge, A.~M. Johnson, M.~Travers, and H.~Choset, ``Design and
  implementation of a three-link brachiation robot with optimal control based
  trajectory tracking controller,'' 2019.

\bibitem{hasegawa1999motion}
Y.~Hasegawa and T.~Fukuda, ``Motion coordination of behavior-based controller
  for brachiation robot,'' in \emph{IEEE SMC'99 Conference Proceedings. 1999
  IEEE International Conference on Systems, Man, and Cybernetics (Cat. No.
  99CH37028)}, vol.~6.\hskip 1em plus 0.5em minus 0.4em\relax IEEE, 1999, pp.
  896--901.

\bibitem{fukuda_brachiator3}
T.~Fukuda, ``A first result of the brachiator iii-a new brachiation robot
  modeled on a siamang,'' \emph{Proc. of the Fifth International Workshop on
  the Synthsis and Simulation of Living Systems}, 1997.

\bibitem{hook-shaped}
Y.~Yamakawa, Y.~Ataka, and M.~Ishikawa, ``{Development of a brachiation robot
  with hook-shaped end effectors and realization of brachiation motion with a
  simple strategy},'' \emph{2016 IEEE International Conference on Robotics and
  Biomimetics, ROBIO 2016}, pp. 737--742, 2016.

\bibitem{farzan-trajopt}
S.~Farzan, A.~P. Hu, E.~Davies, and J.~Rogers, ``{Modeling and control of
  brachiating robots traversing flexible cables},'' \emph{Proceedings - IEEE
  International Conference on Robotics and Automation}, pp. 1645--1652, 2018.

\bibitem{hasegawa2000behavior}
Y.~Hasegawa, Y.~Ito, and T.~Fukuda, ``Behavior coordination and its
  modification on brachiation-type mobile robot,'' in \emph{Proceedings 2000
  ICRA. Millennium Conference. IEEE International Conference on Robotics and
  Automation. Symposia Proceedings (Cat. No. 00CH37065)}, vol.~4.\hskip 1em
  plus 0.5em minus 0.4em\relax IEEE, 2000, pp. 3983--3988.

\bibitem{saito1994learning}
F.~Saito and T.~Fukuda, ``Learning architecture for real robotic
  systems-extension of connectionist q-learning for continuous robot control
  domain,'' in \emph{Proceedings of the 1994 IEEE International Conference on
  Robotics and Automation}.\hskip 1em plus 0.5em minus 0.4em\relax IEEE, 1994,
  pp. 27--32.

\bibitem{fukuda-heuristic}
T.~Fukuda and F.~Saito, ``{Motion control of a brachiation robot},''
  \emph{Robotics and Autonomous Systems}, vol.~18, no. 1-2, pp. 83--93, 1996.

\bibitem{cheng2018deep}
Z.~Cheng, H.~Cheng, and H.~Xu, ``Deep reinforcement learning based brachiation
  control for two-link bio-primate robot,'' in \emph{2018 IEEE International
  Conference on Robotics and Biomimetics (ROBIO)}.\hskip 1em plus 0.5em minus
  0.4em\relax IEEE, 2018, pp. 856--861.

\bibitem{farzan-sos}
S.~Farzan, A.~P. Hu, M.~Bick, and J.~Rogers, ``{Robust control synthesis and
  verification for wire-borne underactuated brachiating robots using
  sum-of-squares optimization},'' \emph{IEEE International Conference on
  Intelligent Robots and Systems}, pp. 7744--7751, 2020.

\bibitem{fukuda1991study}
T.~Fukuda, F.~Saito, and F.~Arai, ``A study on the brachiation type of mobile
  robot (heuristic creation of driving input and control using cmac),'' in
  \emph{Proceedings IROS'91: IEEE/RSJ International Workshop on Intelligent
  Robots and Systems' 91}.\hskip 1em plus 0.5em minus 0.4em\relax IEEE, 1991,
  pp. 478--483.

\bibitem{fukuda-target-dynamics}
J.~Nakanishi, T.~Fukuda, and D.~E. Koditschek, ``{A brachiating robot
  controller},'' \emph{IEEE Transactions on Robotics and Automation}, vol.~16,
  no.~2, pp. 109--123, 2000.

\bibitem{fukuda2013modification}
T.~Fukuda and Y.~Hasegawa, ``Modification on monkey-type mobile robot,''
  \emph{Biologically Inspired Robot Behavior Engineering}, vol. 109, p.~45,
  2013.

\bibitem{hasegawa2001behavior}
Y.~Hasegawa, H.~Tanahashi, and T.~Fukuda, ``Behavior coordination of
  brachiation robot based on behavior phase shift,'' in \emph{Proceedings 2001
  IEEE/RSJ International Conference on Intelligent Robots and Systems.
  Expanding the Societal Role of Robotics in the the Next Millennium (Cat. No.
  01CH37180)}, vol.~1.\hskip 1em plus 0.5em minus 0.4em\relax IEEE, 2001, pp.
  526--531.

\bibitem{nakanishi1999brachiation}
J.~Nakanishi, T.~Fukuda, and D.~E. Koditschek, ``Brachiation on a ladder with
  irregular intervals,'' in \emph{Proceedings 1999 IEEE International
  Conference on Robotics and Automation (Cat. No. 99CH36288C)}, vol.~4.\hskip
  1em plus 0.5em minus 0.4em\relax IEEE, 1999, pp. 2717--2722.

\bibitem{hook-shaped-second}
Y.~Yamakawa, ``Brachiation motion by a 2-dof brachiating robot with hook-shaped
  end effectors,'' \emph{Mechanical Engineering Letters}, vol.~4, pp.
  18--00\,094, 2018.

\bibitem{nakanishi2000leaping}
J.~Nakanishi and T.~Fukuda, ``A leaping maneuvre for a brachiating robot,'' in
  \emph{Proceedings 2000 ICRA. Millennium Conference. IEEE International
  Conference on Robotics and Automation. Symposia Proceedings (Cat. No.
  00CH37065)}, vol.~3.\hskip 1em plus 0.5em minus 0.4em\relax IEEE, 2000, pp.
  2822--2827.

\bibitem{hasegawa1999self}
Y.~Hasegawa, T.~Fukuda, and K.~Shimojima, ``Self-scaling reinforcement learning
  for fuzzy logic controller-applications to motion control of two-link
  brachiation robot,'' \emph{IEEE Transactions on Industrial Electronics},
  vol.~46, no.~6, pp. 1123--1131, 1999.

\bibitem{oliveira-mpc}
V.~M. {De Oliveira} and W.~F. Lages, ``{Linear predictive control of a
  brachiation robot},'' \emph{Canadian Conference on Electrical and Computer
  Engineering}, no. joint 2, pp. 1518--1521, 2006.

\bibitem{oliveira-nmpc}
V.~M. de~Oliveira and W.~F. Lages, ``{Control of a Brachiation Robot With a
  Single Underactuated Joint Using Nonlinear Model Predictive Control},''
  vol.~40, no.~20, pp. 430--435, 2007.

\bibitem{oliveira-realtime}
V.~M. {De Oliveira} and W.~F. Lages, ``{Real-time predictive control of a
  brachiation robot},'' \emph{ETFA 2009 - 2009 IEEE Conference on Emerging
  Technologies and Factory Automation}, 2009.

\bibitem{nakanishi1997preliminary}
J.~Nakanishi, T.~Fukuda, and D.~E. Koditschek, ``Preliminary studies of a
  second generation brachiation robot controller,'' in \emph{Proceedings of
  International Conference on Robotics and Automation}, vol.~3.\hskip 1em plus
  0.5em minus 0.4em\relax IEEE, 1997, pp. 2050--2056.

\bibitem{spong-pfl}
M.~W. Spong, ``Partial feedback linearization of underactuated mechanical
  systems,'' in \emph{Proceedings of IEEE/RSJ International Conference on
  Intelligent Robots and Systems (IROS'94)}, vol.~1.\hskip 1em plus 0.5em minus
  0.4em\relax IEEE, 1994, pp. 314--321.

\bibitem{Wiebe2022}
F.~Wiebe, J.~Babel, S.~Kumar, S.~Vyas, D.~Harnack, M.~Boukheddimi, M.~Popescu,
  and F.~Kirchner, ``Torque-limited simple pendulum: A toolkit for getting
  familiar with control algorithms in underactuated robotics,'' \emph{Journal
  of Open Source Software}, vol.~7, no.~74, p. 3884, 2022.

\bibitem{2022_rss_realaigym}
F.~Wiebe, S.~Vyas, L.~J. Maywald, S.~Kumar, and F.~Kirchner, ``{RealAIGym:
  Education and Research Platform for Studying Athletic Intelligence},'' in
  \emph{Proceedings of Robotics Science and Systems Workshop Mind the Gap:
  Opportunities and Challenges in the Transition Between Research and
  Industry}, New York, July 2022.

\bibitem{2023_wiebe_doublependulum}
F.~Wiebe, S.~Kumar, L.~J. Maywald, S.~Vyas, M.~Javadi, and F.~Kirchner, ``An
  open source dual purpose acrobot and pendubot platform for benchmarking
  control algorithms for underactuated robotics,'' \emph{IEEE Robotics and
  Automation Magazine (RAM)}, 2023, submitted Jan 2023.

\bibitem{Spong1995}
M.~W. Spong, ``{The Swing Up Control Problem For The Acrobot},'' \emph{IEEE
  Control Systems}, vol.~15, no.~1, pp. 49--55, feb 1995.

\bibitem{Betts2010}
J.~T. Betts, \emph{{Practical Methods for Optimal Control and Estimation Using
  Nonlinear Programming}}.\hskip 1em plus 0.5em minus 0.4em\relax Society for
  Industrial and Applied Mathematics, jan 2010.

\bibitem{tedrake2019drake}
R.~Tedrake \emph{et~al.}, ``Drake: Model-based design and verification for
  robotics,'' \emph{URL https://drake. mit. edu}, 2019.

\bibitem{gill2005snopt}
P.~E. Gill, W.~Murray, and M.~A. Saunders, ``Snopt: An sqp algorithm for
  large-scale constrained optimization,'' \emph{SIAM review}, vol.~47, no.~1,
  pp. 99--131, 2005.

\bibitem{todorov2012mujoco}
E.~Todorov, T.~Erez, and Y.~Tassa, ``Mujoco: A physics engine for model-based
  control,'' in \emph{2012 IEEE/RSJ International Conference on Intelligent
  Robots and Systems}.\hskip 1em plus 0.5em minus 0.4em\relax IEEE, 2012, pp.
  5026--5033.

\bibitem{schulman2017proximal}
J.~Schulman, F.~Wolski, P.~Dhariwal, A.~Radford, and O.~Klimov, ``Proximal
  policy optimization algorithms,'' 2017.

\bibitem{stable-baselines3}
A.~Raffin, A.~Hill, A.~Gleave, A.~Kanervisto, M.~Ernestus, and N.~Dormann,
  ``Stable-baselines3: Reliable reinforcement learning implementations,''
  \emph{Journal of Machine Learning Research}, vol.~22, no. 268, pp. 1--8,
  2021.

\bibitem{underactuated}
\BIBentryALTinterwordspacing
R.~Tedrake, \emph{Underactuated Robotics}, 2022. [Online]. Available:
  \url{https://underactuated.csail.mit.edu/}
\BIBentrySTDinterwordspacing

\bibitem{bertsekas2012dynamic}
D.~Bertsekas, \emph{Dynamic programming and optimal control: Volume I}.\hskip
  1em plus 0.5em minus 0.4em\relax Athena scientific, 2012, vol.~1.

\bibitem{kaspar2020sim2real}
M.~Kaspar, J.~D.~M. Osorio, and J.~Bock, ``Sim2real transfer for reinforcement
  learning without dynamics randomization,'' in \emph{2020 IEEE/RSJ
  International Conference on Intelligent Robots and Systems (IROS)}.\hskip 1em
  plus 0.5em minus 0.4em\relax IEEE, 2020, pp. 4383--4388.

\end{thebibliography}
\end{samepage}

\end{document}